\documentclass{article} %
\usepackage{iclr2024_conference,times}

\usepackage{amsmath,amsfonts,bm}

\def\eqref#1{equation~\ref{#1}}

\def\1{\bm{1}}

\def\rmB{{\mathbf{B}}}

\def\rmF{{\mathbf{F}}}

\def\rmS{{\mathbf{S}}}

\def\mW{{\bm{W}}}

\DeclareMathAlphabet{\mathsfit}{\encodingdefault}{\sfdefault}{m}{sl}
\SetMathAlphabet{\mathsfit}{bold}{\encodingdefault}{\sfdefault}{bx}{n}

\usepackage[utf8]{inputenc} %
\usepackage[T1]{fontenc}    %
\usepackage{url}            %
\usepackage{booktabs}       %
\usepackage{amsfonts}       %
\usepackage{nicefrac}       %
\usepackage{microtype}      %
\usepackage{colortbl}
\usepackage{amsthm}
\usepackage{thmtools}
\usepackage{bbm}
\usepackage{mathtools}
\usepackage{algorithm}
\usepackage{algorithmic}
\usepackage{multirow}
\usepackage{hhline}
\usepackage{xcolor,colortbl}
\definecolor{refcolor}{rgb}{0.2549, 0.4117, 0.8823}
\usepackage[allcolors=refcolor,colorlinks=true]{hyperref}

\usepackage{caption}
\usepackage{subcaption}
\usepackage{wrapfig}
\usepackage{enumitem,kantlipsum}
\usepackage{tabularx}
\usepackage{amssymb}
\usepackage[textsize=meduim]{todonotes}

\title{
TAIL: Task-specific Adapters for Imitation Learning with Large Pretrained Models
}

\iclrfinalcopy

\author{%
  Zuxin Liu$^{1}$, Jesse Zhang$^{2}$, Kavosh Asadi$^3$, Yao Liu$^3$, Ding Zhao$^1$, Shoham Sabach$^3$, Rasool Fakoor$^3$ \\
  $^1$Carnegie Mellon University, $^2$ University of Southern California, $^3$ Amazon Web Services\\
}

\usepackage{scalerel}

\definecolor{pltcolor}{rgb}{0.18039, 0.45882, 0.713725}
\newcommand{\iclrnew}[1]{\textcolor{black}{#1}}
\newcommand*{\mybox}[1][pltcolor]{\scalerel*{\textcolor{#1}{\rule{1ex}{1ex}}}{d}}

\newcommand{\method}{TAIL}
\newcommand{\methodlong}{TAIL (\textbf{T}ask-specific \textbf{A}dapters for \textbf{I}mitation \textbf{L}earning)}

\begin{document}

\maketitle

\begin{abstract}
The full potential of large pretrained models remains largely untapped in control domains like robotics. 
This is mainly due to data scarcity and computational challenges associated with training or fine-tuning large models for such applications. 
Prior work mainly emphasizes either effective \emph{pretraining} of large models for decision-making or \iclrnew{single-task adaptation. But real-world problems will require data-efficient, \emph{continual adaptation} for new control tasks.} 
Recognizing these constraints, we introduce \methodlong, a framework for efficient adaptation to a stream of new control tasks.
Inspired by recent advancements in parameter-efficient fine-tuning in language domains, we explore efficient fine-tuning techniques---e.g., Bottleneck Adapters, P-Tuning, and Low-Rank Adaptation (LoRA)---in \method\ to adapt large pretrained models for new tasks with limited demonstration data. 
Our extensive experiments comparing prevalent parameter-efficient fine-tuning techniques and adaptation baselines suggest that \method\ with LoRA can achieve the best post-adaptation performance with only 1\% of the trainable parameters of full fine-tuning while avoiding catastrophic forgetting and preserving adaptation plasticity in continual learning settings.

\end{abstract}

\section{Introduction}

A desired property of an autonomous agent is the ability to adapt efficiently to novel tasks. In vision and language domains, large pretrained models have demonstrated adaptation to new tasks with just a few examples through prior knowledge obtained from internet-scale datasets~\citep{brown2020language,radford2021learning,touvron2023llama}. 
Similar methods have also been applied in decision-making and control applications~\citep{brohan2022rt,driess2023palme,rt22023arxiv}.
However, new control tasks are more difficult to adapt to than the aforementioned vision and language domains due to (1) the lack of internet-scale control data and (2) how optimal actions can vary significantly from task-to-task, even under shared observation spaces. 
As such, these large-scale decision-making models still rely on a close alignment between training and testing tasks.

In contrast, agents deployed in challenging environments need to adapt to major task variations---take, for example, a general household robot. 
Equipped with a factory-pretrained policy, the robot will be employed in unique ways by every household. Thus, the robot will need to \emph{continually adapt} in order to best serve each one, e.g., by fine-tuning its capabilities on a few demonstrations~\citep{actionablemodels2021arxiv,awopt2021corl,mtopt2021arxiv,chen2021context, yao2024constraint}.
Because most prior decision-making papers adapt to new tasks by fine-tuning the entire model~\citep{ gupta2022dbap,bousmalis2023robocat,zhang2023sprint,zhang2023bootstrap,open_x_embodiment_rt_x_2023, liu2023constrained}, 
mastering each new skill requires great computational cost and often leads to catastrophic forgetting of old ones. An alternative approach would be to store a separate policy per new task, which leads to unreasonable storage requirements.
\iclrnew{Some prior work investigates efficient adaptation of large models to a single task suite~\citep{liang2022transformer, schmied2023learning, sharma2023lossless}, but this realistic continual learning setting brings out additional problems to consider, warranting further investigation.}
What would be the best way for agents to \emph{efficiently adapt} to a stream of novel tasks without having to trade off computation, storage, and performance on older tasks?

To answer this question, we propose \textbf{T}ask-specific \textbf{A}dapters for \textbf{I}mitation \textbf{L}earning, shown in Fig. \ref{fig:model-arch}, a framework for efficient adaptation to new control tasks.
Through \method\ we (1) effectively incorporate lightweight adapter modules into pretrained decision-making models and (2) comprehensively compare efficient adaptation techniques implemented in \method\ in a continual imitation learning setting.
Notably, we examine parameter-efficient adaptation techniques (PEFT) used for large language models; we explore the potential of adapters~\citep{houlsby2019parameter}, prefix tuning~\citep{li2021prefix}, and low-rank adaptation (LoRA)~\citep{hu2021lora} in fostering efficient and continual adaptation in large pretrained decision-making models.
These works stand out as they introduce a small number of \emph{new} parameters which help: avoid catastrophic forgetting, maintain training plasticity for continual learning, avoid overfitting with limited adaptation data, and reduce computational and memory burden.
Investigating these works in control tasks for a realistic continual learning setup specifically is important because, unlike in language domains, test task losses are often not proportional to test task performance~\citep{ross2011dagger, ke2020imitation}---efficient adaptation insights from language models may not transfer to decision-making ones.
Thus, independent investigation of these adaptation techniques for decision-making is crucial for deploying continually adapting agents in the real world.

We compare PEFT techniques implemented in \method\ against commonly used adaptation methods in the imitation learning literature. 
In our experiments, we discover that \method\ with LoRA leads to the best post-adaptation performance as %
it preserves the original pretrained representations while being resilient against overfitting in the limited-data regime. 
These capabilities are especially important for agents operating in new, challenging environments, such as the aforementioned household robots. 
Our analysis also reveals important insights into the strengths and limitations of each adaptation strategy.
Instead of performing full fine-tuning of the entire model, \method\ only introduces a small number of additional parameters without making changes to the original model. These additional parameters make up a mere $\boldsymbol{1.17\%}$ of the size of the original model. Importantly, this results in approximately $\boldsymbol{23\%}$ less GPU memory consumption to achieve $\boldsymbol{22\%}$ higher forward adaptation success rate than full fine-tuning while avoiding catastrophic forgetting. Notably, these results are contrary to many results from the vision and language model literature which show that full fine-tuning works better \citep{he2022towards, mao-etal-2022-unipelt, chen2022adaptformer, schmied2023learning}.

In summary, this work bridges a crucial gap in research into efficient and continual adaptation for pretrained decision models by introducing a framework for continual imitation learning, \method, and thoroughly analyzing the effects of different efficient adaptation methods. \iclrnew{Comprehensive experiments demonstrate that \method\ outperforms standard continual learning and prior single-task adaptation baselines.} %

\vspace{-4mm}
\section{Related Work}
\vspace{-3mm}
\textbf{Pretrained Models for Control.}
Researchers have long studied the use of pretrained models for better downstream transfer to related tasks~\citep{bozinovski1976influence, schmidhuber1992learning, dietterich1997special}.
Recent works have examined using the representations learned by pretrained visual models for control~\citep{shridhar2022cliport, nair2022r3m, ma2022vip, ma2023liv, vc2023}. 
These methods leverage representations acquired from large task-agnostic datasets, such as Ego4D~\citep{grauman2022ego4d}, or through self-supervised objectives. 
However, there's evidence that simply utilizing these pretrained features may not be as useful for downstream task performance \citep{hansen2022on}. 
Meanwhile, another recent line of work directly trains large pretrained models for control~\citep{brohan2022rt, reed2022gato, driess2023palme, jiang2023vima, rt22023arxiv, bousmalis2023robocat}. 
These methods either do not attempt adaptation to new tasks, or perform expensive full-fine-tuning for adaptation. 
In contrast, our method, \method, is a framework for efficient adaptation of decision-making models, like the aforementioned large pretrained control models, and investigates ways to adapt such models efficiently to multiple new tasks.

\textbf{Parameter-Efficient Fine-Tuning (PEFT).}
PEFT has gained traction as a way to adapt pretrained models without significantly increasing parameters. \iclrnew{\citet{rebuffi2018} demonstrated that residual adapters for smaller, CNN-based vision models are effective in non-control supervised learning settings.} More recently, transformer-focused techniques such as transformer adapter modules~\citep{houlsby2019parameter}, LoRA~\citep{hu2021lora}, and prompt tuning~\citep{li2021prefix} incorporate lightweight modules or prompts optimized for downstream tasks, all while preserving the original model weights. PEFT offers several advantages over full fine-tuning: it's faster, less susceptible to overfitting, retains prior capabilities, and facilitates efficient task-switching. While PEFT has been successful in both language and vision domains~\citep{chen2022adaptformer,schmied2023learning}, its continuous adaptation for large decision-making models is not yet thoroughly examined. 
\citet{liang2022transformer, sharma2023lossless}, \citet{xu2022prompting}, and \citet{xu2023hyper} propose the use of adapters, prompt-tuning, and hyper-network in robotics settings, but they do not examine other PEFT methods and focus on adaptation to a single task suite. We instead examine the performance of various state-of-the-art PEFT techniques implemented with \method\ in the \emph{continual learning} scenario.

\textbf{Continual Learning.}
Continual learning in control~\citep{Thrun95,McCloskey1989CatastrophicII, fu2022model} is a long-studied problem with applications to many real-world situations.
In general, agents should be able to transfer knowledge (e.g., by continually fine-tuning) or experience (e.g., training data) from previously learned tasks to new tasks~\citep{lopez2017gradient,Traore19DisCoRL,fakoor2019meta,caccia2023taskagnostic}.
However, with large pretrained models trained on large datasets, fine-tuning the entire model is computationally costly yet risks catastrophic forgetting, and transferring training data from other tasks is too memory inefficient in the face of a large stream of new tasks.
Therefore, we present a study into efficient fine-tuning techniques which, when integrated with \method, can help inform future research of continual learning.

\vspace{-3mm}
\section{Preliminaries}
\vspace{-2mm}
In this section, we introduce our problem setting (Sec.~\ref{sec:prelim:problem formulation}), review large, pretrained models for decision-making (Sec.~\ref{sec:prelim:robot fms}), and discuss traditional adaptation methods in this area (Sec.~\ref{sec:prelim:adapting}).

\begin{figure}[t]
\centering
\includegraphics[width=0.90\linewidth]{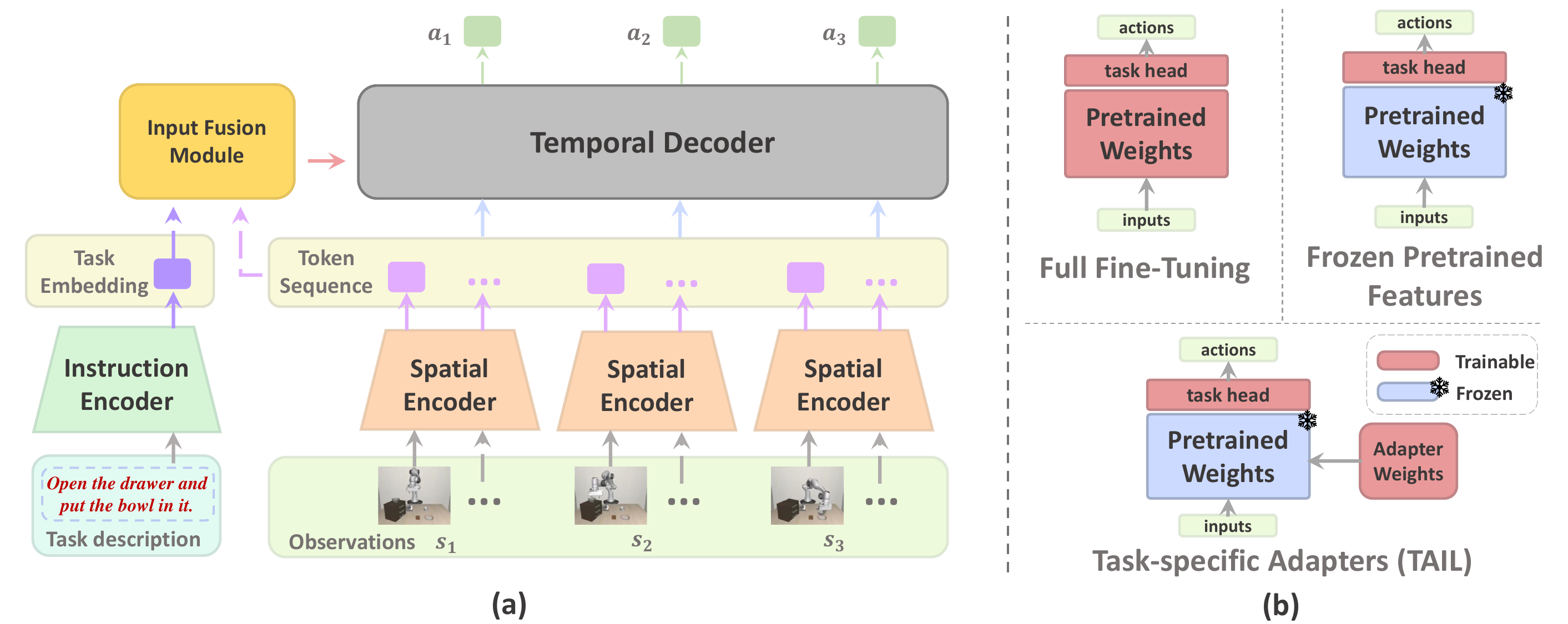}
\vspace{-2mm}
\caption{ \small \textbf{(a)}: The multi-modal, transformer policy architecture we utilize for pretraining. We encode language task descriptions with a pretrained CLIP \textcolor{teal}{instruction encoder} and image observations with a pretrained CLIP \textcolor[HTML]{F7921D}{spatial encoder}. We additionally encode state observations (not pictured) which, along with the observation embeddings, are embedded into a sequence of tokens used by the \textcolor{gray}{temporal decoder} transformer to predict single-step action distributions. We include an \textcolor[HTML]{FDBC42}{input fusion module} to explicitly combine the task embedding with the observation token sequence for better instruction-following ability.
\textbf{(b)}: The three types of fine-tuning paradigms we test, with \method\ at the bottom right. For further architecture details, see Appendix Sec.~\ref{sec:appendix:architecture}.}
\label{fig:model-arch}
\vspace{-4mm}
\end{figure}

\vspace{-3mm}
\subsection{Continual Imitation Learning}
\vspace{-2mm}
\label{sec:prelim:problem formulation}

The agent encounters a sequence of \(K\) tasks, denoted as \(\{\mathcal{T}_1, \ldots, \mathcal{T}_K\}\). Each task \(\mathcal{T}_k = (\mu_k^0, g_k)\) is characterized by an initial state distribution \(\mu_k^0\) and a goal predicate \(g_k\). Goals for tasks can be specified using language instructions, providing clear context~\citep{jang2021bcz, zhang2023sprint}.
For every task \(\mathcal{T}_k\), the agent receives \(N\) demonstration trajectories \(\mathcal{D}_k = \{\tau_{k}^{1}, \ldots, \tau_{k}^{N}\}\). 
In this paper, we use the standard behavioral cloning loss to optimize the agent's policy $\pi$ over these demonstrations, however we note that \method\ can be used with other training objectives as well:

 \begin{equation}
\hat{\boldsymbol{\theta}} = \min_{\boldsymbol{\theta}} \sum_{k=1}^K 
\underset{s_t, a_t \sim \mathcal{D}_k}{\mathbb{E}}
\left[\sum_{t=0}^{l_k} \mathcal{L}\left(\pi(a|s_{\leq t}, \mathcal{T}_k;\boldsymbol{\theta}), a_k^t\right)\right].
\label{eq:bc-loss}
\end{equation}
Here, $\mathcal{L}$ is a supervised action prediction (e.g., mean squared error or negative log likelihood) loss, \(l_k\) is the length of demonstrations for task \(\mathcal{T}_k\), and $\bm{\theta}$ refers to the \textit{learnable parameters} of the network.
Notably, after learning task \(\mathcal{T}_k\), the agent cannot access \emph{additional} data from preceding tasks. %
This presents a continual learning challenge, emphasizing the importance of transferring knowledge across tasks without the risk of catastrophic forgetting~\citep{McCloskey1989CatastrophicII}. 

\vspace{-2mm}
\subsection{Pretrained Decision-Making Models}
\vspace{-1mm}
\label{sec:prelim:robot fms}

Here, we briefly describe common features of large pretrained decision-making model architectures used for embodied agents.
We incorporate key components shared amongst these models into the architecture of the model that we pretrain to evaluate efficient adaptation, pictured in Fig.~\ref{fig:model-arch}(a).

\textbf{Transformer Backbone.}
Most recent work training large-scale decision-making models~\citep{brohan2022rt, shafiullah2022behavior, rt22023arxiv} utilize a transformer backbone ~\citep{vaswani2017attention} that attends to tokenized observations from prior timesteps. 
We adopt a standard GPT-2~\citep{radford2019GPT2} transformer decoder (Fig.~\ref{fig:model-arch}(a), \textcolor{gray}{temporal decoder}) with separate encoders for each input modality and continuous action distribution outputs.

\textbf{Pretrained Input Encoders.}
Encoders pretrained on large, diverse datasets can produce rich, well-structured embeddings which make it easier to learn the downstream tasks~\citep{jang2021bcz, brohan2022rt}.
Therefore, we utilize pretrained CLIP image and textual encoders~\citep{radford2021learning}.

\textbf{Input Modality Fusion.}
The idea of explicitly ``fusing'' different input modalities has seen great success not only in domains like vision and language~\citep{perez2017film}, but also in agent learning~\citep{jang2021bcz, brohan2022rt}. 
Similarly, we utilize FiLM layers~\citep{perez2017film} (Fig.~\ref{fig:model-arch}(a), \textcolor[HTML]{FDBC42}{input fusion module}) to fuse language task specifications with observations.%

\vspace{-2mm}
\subsection{Adapting pretrained models for new tasks}
\vspace{-2mm}
\label{sec:prelim:adapting}
One standard adaptation method in prior research is full fine-tuning (FFT) of all model parameters (Fig~\ref{fig:model-arch}(b), top left). 
Though straightforward, it is resource-intensive and prone to overfitting with limited data~\citep{bousmalis2023robocat}.
There is also a risk of distorting pretrained features, resulting in the loss of prior tasks---a phenomenon known as \textbf{catastrophic forgetting}~\citep{McCloskey1989CatastrophicII}.
Evidence also suggests that extensive fine-tuning might undermine a model's rapid adaptability to new tasks, an effect referred to as the loss of \textbf{model plasticity and capacity} \citep{kumar2022finetuning, lyle2022understanding, kumar2023maintaining}. 
Such issues become more prominent in continual learning contexts~\citep{lopez2017gradient}. Moreover, duplicating a sizable model for each subsequent task is neither efficient nor practical due to storage limitations.

Another standard adaptation method is the use of frozen pretrained features (FPF, Fig~\ref{fig:model-arch}(b) top right). FPF ensures the retention of knowledge acquired from previous tasks by tuning a task-specific head. 
However, as noted in \citet{sharma2023lossless}, it is not expressive enough for out-of-distribution or especially complex tasks.
Given these challenges, there's a clear need for a more advanced fine-tuning paradigm that addresses catastrophic forgetting while maintaining model plasticity for adapting to new tasks, all in a data and computationally resource-efficient manner.%

\vspace{-3mm}
\section{Task-specific adapters for imitation learning}
\vspace{-2mm}
In this section, we outline how we perform efficient adaptation on pretrained models through our \textbf{T}ask-specific \textbf{A}dapters for \textbf{I}mitation \textbf{L}earning framework, depicted in Fig \ref{fig:model-arch}(b).
Different from the FPF approach which simply substitutes the policy head for every new task, \method\ introduces a small set of new weights, serving as a lightweight plugin to address specific tasks. 
This concept draws inspiration from parameter-efficient adaptation techniques prevalent in the language model area. These methods offer several advantages as they:
(1) add a few parameters (typically between $0.1\% \sim 2\%$) to preserve the original features, thereby enhancing model plasticity for continual learning and avoiding catastrophic forgetting \citep{kumar2023maintaining},
(2) are resilient to overfitting when adaptation data is scarce,
(3) are more computationally and storage-efficient than FFT.

Next, we delve into three prominent weight integration techniques for Transformer-based pretrained models in Sec.~\ref{sec:adapter-integration}, followed by a case study illustrating the application of this framework in continual imitation learning scenarios in Sec.~\ref{sec:tail-continual-learning}.

\vspace{-2mm}
\subsection{Adapter Weights Integration}
\begin{figure}[t]
    \centering
    \includegraphics[width=0.86\linewidth]{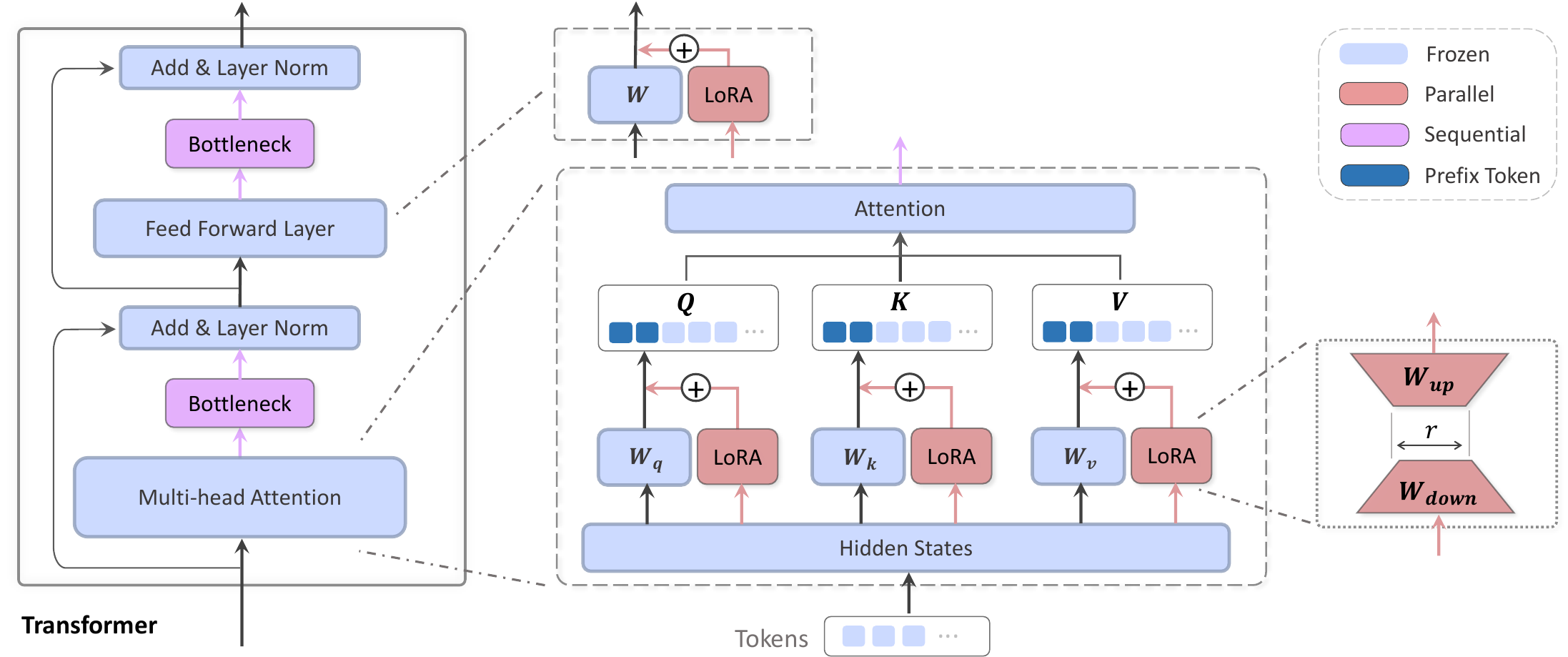}
    \caption{Demonstration of three weight integration styles of \method\ for a Transformer block: \textcolor[HTML]{E5ADFF}{sequential} (bottleneck adapter), \textcolor[HTML]{E99999}{parallel} (LoRA), and \textcolor[HTML]{9DC3E6}{prefix token} (prefix/prompt-tuning).}
    \label{fig:integration}
    \vspace{-4mm}
\end{figure}

\vspace{-1mm}
\label{sec:adapter-integration}
The concept of an adapter can be best conceptualized as a modular plugin to the base model, customized for specific downstream tasks, that does not affect the model's pretrained representations. 
We mainly explore three prevalent styles of integration for \method: \textbf{Parallel}~\citep{hu2021lora}, \textbf{Sequential}~\citep{houlsby2019parameter, sharma2023lossless}, and \textbf{Prefix Token}~\citep{li2021prefix, lester2021power, liu2023gpt}, all of which are showcased with a Transformer block in Fig.~\ref{fig:integration}.
Parallel and sequential integration techniques are generally applicable to any model with feedforward layers, while the prefix token style method is especially tailored for Transformers.

Given a pretrained model, let's consider \emph{one} layer weight matrix in it, denoted as $\mW \in \mathbb{R}^{d \times k}$. Its input and output hidden states are $ h_{in} \in \mathbb{R}^{d}$ and $ h_{out} \in \mathbb{R}^{k}$, respectively. We have $ h_{out} = \mW^\top h_{in}$.
Next, we detail how to apply parallel and sequential insertions to the pretrained weight matrix.

\textbf{Parallel Integration (LoRA).} This integration method, often associated with Low-Rank Adaptation (LoRA) \citep{hu2021lora}, introduces trainable low-rank matrices $\mW_{down} \in \mathbb{R}^{d \times r}$ and \( \mW_{up} \in \mathbb{R}^{r \times k}\). Here, $r \ll \min(d, k)$ represents the rank and is usually much smaller than the dimensions of the original matrix.
These matrices are typically integrated in parallel with the original weight matrix $\mW$ through addition, as shown as \textcolor[HTML]{E99999}{LoRA} in  {Fig. \ref{fig:integration}:
\begin{equation}
     h_{out} = \mW^\top h_{in} + \alpha \mW_{up}^\top\mW_{down}^\top h_{in},
\end{equation}
with \( \alpha \) being a hyperparameter to modulate task-specific adjustments. The above equation can also be formulated as:
$
     h_{out} 
    = (\mW + \alpha \mW_{down}\mW_{up})^\top  h_{in} = (\mW + \alpha \Delta \mW)^\top  h_{in}
$,
where $\Delta \mW$ denotes the weight modifications for new tasks, and thus the columns of $\mW_{down}$ and $\mW_{up}$ can be interpreted as a new basis that contains task-specific knowledge.
As observed by \citet{aghajanyan2020intrinsic}, despite projecting to a condensed subspace with small “intrinsic dimensions,” pretrained models can still learn effectively.
By introducing the two low-rank matrices, the original weight matrices $\mW$ can be adeptly tailored with a minimal increase in parameters. Though LoRA was originally crafted for large language models—specifically for the query and value projections matrices \( W_Q \) and \( W_V \) in multi-head attention \citep{hu2021lora}---it is easily applied to other linear layers as well, such as the Transformer's feedforward layers \citep{chen2022adaptformer}.

\textbf{Sequential Integration (Bottleneck Adapter).} Renowned in the language model domain, the Bottleneck Adapter introduces bottleneck layers within the model \citep{houlsby2019parameter, sharma2023lossless} by appending a trainable bottleneck layer after the feedforward network in each Transformer layer. Similar to LoRA, this bottleneck consists of down and up projections, $\mW_{down}$ and $\mW_{up}$, which first shrink then restore the dimensions of token hidden states. 
Formally, for the feedforward network's input \(  h_{in} \) and a bottleneck size \( r \), the output \(  h_{out} \) is:
\begin{equation}
  h_{out} = \mW_{up}^\top\phi\left(\mW_{down}^\top (\mW^\top  h_{in})\right),
\end{equation}
where $ \phi $ denotes a nonlinear activation function. The \textcolor[HTML]{E5ADFF}{Bottleneck Adapter} (Fig.~\ref{fig:integration}) acts as a filter, isolating relevant information for specific tasks. Yet, filtering often requires a larger bottleneck size compared to that of LoRA, leading to more parameters. Additionally, the sequential insertion can increase latency compared to the parallel nature of LoRA \citep{hu2021lora}.
    
\textbf{Prefix Token Integration (Prefix \& Prompt-Tuning).} In this style, a set of learnable prefix tokens are appended or prepended to the input sequence \citep{li2021prefix, lester2021power,liu2023gpt}. 
Let's consider an input sequence \( \mathbf{s} \in \mathbb{R}^{n \times d} \), where \( n \) is the sequence length and \( d \) is the embedding dimension. The prefix tokens can be represented as \( \mathbf{p} \in \mathbb{R}^{m \times d} \), where \( m \) denotes the number of prefix tokens. 
These vectors act like virtual tokens which the original tokens can attend to. They are initialized and learned during the task-specific adaptation phase.
The modified input sequence, after appending the prefix tokens, can be expressed as \( \mathbf{S} = [\mathbf{p}; \mathbf{s}] \in \mathbb{R}^{(m+n) \times d} \). 
The model then processes this extended sequence. 
These prefix tokens can be viewed as task descriptors that are designed to guide the model towards the desired task-specific behavior
(see {\large\mybox{}} in Fig.~\ref{fig:integration}). 

With adapters, we can treat the optimization from Eq.~\ref{eq:bc-loss} as one over adapter weights instead, where the model is parametrized by $\hat{\bm{\theta}} = \{ \bm{\theta}, \bm{\omega} \}$ and $\bm{\omega}$ is the set of adapter weights we are optimizing for.

\vspace{-2mm}
\subsection{\method\ for continual imitation learning}
\vspace{-1mm}
\label{sec:tail-continual-learning}

We consider the continual imitation learning problem as a typical application of the proposed \method\ adaptation paradigm. 
The goal of continual imitation learning is to ensure that the model performs effectively on the current task and without significant degradation of performance in past tasks. \\
Given pretrained model weights, denoted as $\bm{\theta}$, and a new task $\mathcal{T}_k$ with demonstrations $\mathcal{D}_k = \{\tau_{k}^{1}, \ldots, \tau_{k}^{N}\}$, we initialize the task-specific adapter weight $\bm{\omega}_k$ with far less parameters than the base model: $|\bm{\omega}_k| \lll |\bm{\theta}|$. The adapter weights are inserted into the model through the integration methods introduced in Sec.~\ref{sec:adapter-integration}. By optimizing the behavior cloning loss in Eq.~\ref{eq:bc-loss} w.r.t $\bm{\omega}_k$ while keeping the pretrained weights frozen, the policy adapts to $\mathcal{T}_k$ without interfering with previous tasks.

To execute a task, the corresponding lightweight adapters are loaded as a plugin of the pretrained network weights. For example, when revisiting a prior task $T_j$, where $j \leq k$, the model is configured to solely activate the $j$-th adapter $\bm{\omega}_j$. %
This entire procedure can be streamlined as follows:
\begin{enumerate}[leftmargin=0.8cm]
    \item For an incoming task \( \mathcal{T}_k \), acquire the training set \( \mathcal{D}_k \), initialize a task-specific adapter $\bm{\omega}_k$.
    \item Combine adapter $\bm{\omega}_k$ with the base model $\bm{\theta}$ using either parallel, sequential, or prefix token.
    \item Train the adapter on \( \mathcal{D}_k \) to optimize Eq.~\ref{eq:bc-loss} for $\bm{\omega}_k$, keeping pretrained parameters $\bm{\theta}$ frozen. 
\end{enumerate}
In essence, \method\ ensures task-specific knowledge is contained within the adapters, thereby enabling efficient adaptation without catastrophic forgetting.
It's also worth noting that the TAIL framework is flexible. 
The choice of integration method or the specific architecture of the adapter can be tailored based on the complexity of the task or the available computational resources.

\vspace{-3mm}
\section{Experiments}
\vspace{-2mm}
\label{sec:exp}
In this section, we evaluate \method\ on a wide range of tasks and benchmark its performance against other fine-tuning approaches.
We mainly aim to answer the following questions: 
(1) Which efficient adaptation methods in \method\ work best? 
(2) Can \method\ prevent catastrophic forgetting of previously learned tasks, while allowing more efficient forward adaptation to new tasks over standard adaptation methods? 
(3) What are the computational efficiencies gained by using \method?
Addressing them requires a set of diverse tasks in realistic environments, as we describe in the following section.

\vspace{-3mm}
\subsection{Datasets and Benchmark Suites}
\vspace{-2mm}

\begin{figure}[t]
\vspace{-5mm}
\centering
\includegraphics[width=0.78\textwidth]{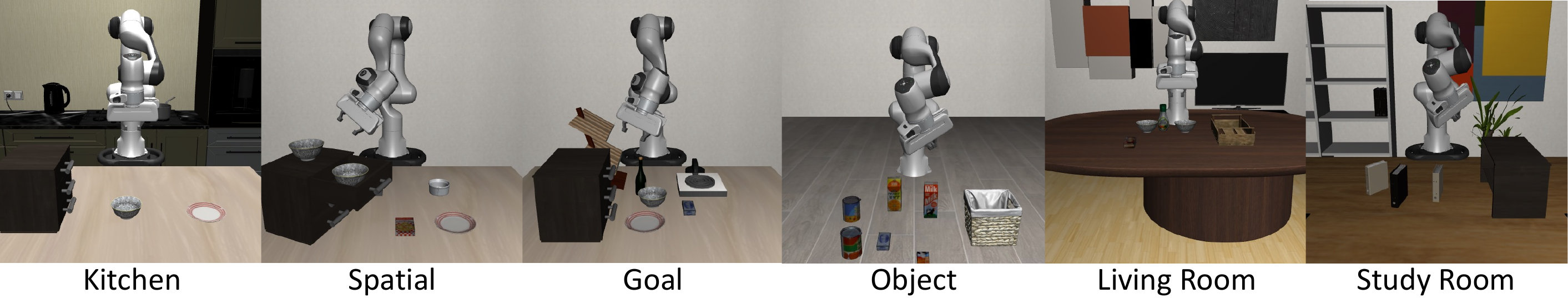}
\caption{\small Our task suites for continual imitation learning (excluding LIBERO-10). The robot, placed in a tabletop environment, is equipped with a 6-DOF arm and a parallel gripper. It receives RGB images from two views, joint states, 
and language instructions, and is tasked with producing continuous actions to control its arm. }
\label{fig:libero_demo}
\vspace{-5mm}
\end{figure}

We utilize the LIBERO robotic manipulation continual learning benchmark~\citep{liu2023libero}, which features a diverse range of tasks that mirror human daily activities, such as turning on a stove, moving books, and opening drawers. 
Each task is specified via natural language instructions, for instance, \textit{"Open the top drawer of the cabinet, and put the bowl in it."} 

We craft a \textit{pretraining} task suite, named \textbf{Kitchen}, involving 40 diverse tasks sourced from the LIBERO-90 dataset's kitchen scenes. We then evaluate \textit{adaptation} to 5 separate task suites. LIBERO contains 3 task suites tailored for continual learning, focusing on evaluating different aspects of knowledge adaptation: 
the \textbf{Spatial} task contains the same objects in each scene but with different spatial layouts; each task in the \textbf{Goal} suite has distinct goals (such as open the drawer, or turn on the stove), while keeping the objects and layout fixed; the \textbf{Object} suite contains pick-and-place tasks for different objects in the scene but with the same layout. To create a more comprehensive experimental setting, we also create 2 \emph{additional} task suites (from LIBERO-90): \textbf{Living Room}, and \textbf{Study Room}.
We adopt 8 tasks from each of the 5 adaptation task suites, respectively.
Finally, we separately evaluate each task sequentially in \textbf{LIBERO-10}, a benchmark with 10 challenging long-horizon tasks. 
See Fig.~\ref{fig:libero_demo} for task suite examples and Appendix Sec.~\ref{sec:appendix:task details} for more details.

\vspace{-3mm}
\subsection{Experiment setup}
\vspace{-2mm}
\textbf{Evaluation metrics.}
The primary metric we report is average per-task \emph{suite} success rate, measured by checking if current state aligns with pre-defined goal states.
For continual learning, we also assess \textbf{Forward Transfer} (FWT) and \textbf{Backward Transfer} (BWT) across the curriculum of suites. Following the metric proposed in LIBERO~\citep{liu2023libero}, 
FWT is computed by the maximum success rate one algorithm can achieve when adapting to a new task.
We denote FWT at task $k$ as $\rmF_k$.
Meanwhile, BWT measures the success rate increase on previous tasks.
Namely, when adapting to the $k$-th task, we record the best FWT model on this task and then evaluate this model on all previous $k-1$ tasks, obtaining success rate $\rmS_{i}, 1 \leq i \leq k-1$. Then we compute the success rate difference between the new model and the best FWT of the previous $k-1$ tasks and then average among them to obtain the BWT metric: \iclrnew{$\rmB_k = \frac{1}{k-1} \sum _{i=1}^{k-1} (\rmS_{i} - \rmF_{i})$}. 
For both metrics, higher is better.

\begin{figure}[t]
\centering
\includegraphics[width=0.95\textwidth]{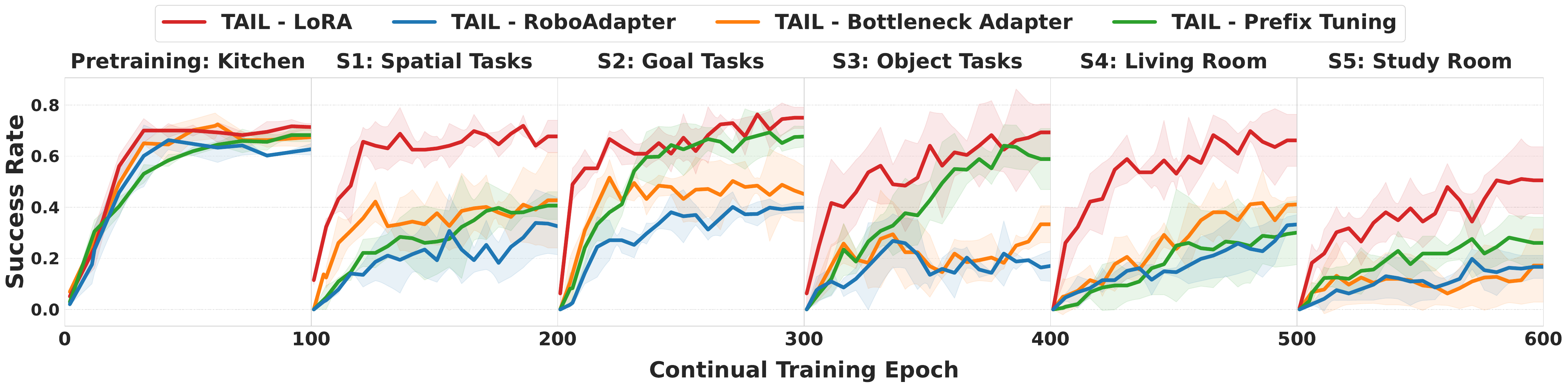}
\vspace{-2mm}
\caption{\small Success rates for different types of adapters under our \method\ framework. None of these methods suffer from catastrophic forgetting, so backward evaluation results are not presented here.
\textcolor{red}{LoRA} performs best across all tasks, 
underscoring the benefits of the parallel integration approach.}
\label{fig:training_curve_adapter}
\vspace{-5mm}
\end{figure}

\textbf{Model architecture.}
We adopt the CLIP-base model \citep{radford2021learning} as both the spatial encoder and the language instruction encoder, each with 12 transformer layers. A 6-layer GPT2 structure \citep{radford2019language} serves as our temporal encoder, with the FiLM module \citep{perez2017film} handling input fusion. These components are well-regarded in the literature \citep{chen2021decision, brohan2022rt, jiang2023vima}. 
Further architectural details can be found in Appendix~\ref{sec:appendix:architecture}.

\textbf{Continual Learning Baselines.}
We adopt four baselines: Full Fine-Tuning (FFT), Frozen Pretrained Features (FPF) which mirrors the linear probing method~\citep{kumar2022finetuning} but also tunes both the policy head and fusion module \emph{per task}, Experience Replay (ER) \citep{chaudhry2019tiny} which uses a 50-50 data split between new and previous task data while adapting to a new task~\citep{rolnick2019experience}, Elastic Weight Consolidation (EWC)~\citep{kirkpatrick2017overcoming} which regularizes updates of crucial parameters from earlier tasks based on their Fisher information, and \iclrnew{PackNet~\citep{packnet2018} which prunes parameters to then be re-learned for every new task}. \iclrnew{These all use the same model and task conditioning, i.e., language, as \method.} 
Further baseline details in Appendix \ref{sec:appendix:baselines}. 

\textbf{\method{}\ Adapters.} 
We use LoRA~\citep{hu2021lora}, Bottleneck Adapter~\citep{houlsby2019parameter}, and Prefix Tuning~\citep{li2021prefix} to represent parallel, sequential, and prefix integration styles. RoboAdapter~\citep{sharma2023lossless}, a specific implementation for decision-making, stands as another \emph{sequential} integration style. Unlike the Bottleneck Adapter that applies weights at every transformer layer, it introduces weights only at specific transformer layers and exclusively after the feedforward layer. Configuration specifics and more details for these adapters are available in Appendix \ref{sec:appendix:tail_adapters}.

\textbf{Training, Adaptation, and Evaluation.}
Each task provides 50 \iclrnew{successful} human demonstrations. These are divided into 40 training trajectories and 10 for validation. \textit{\iclrnew{We report success rates over 10 scenes with initial states that are unseen in training}}. This \iclrnew{limited demonstration} setup offers an opportunity to determine which technique is less prone to overfitting in data-restricted conditions. 
Given our focus on evaluating the adaptation of large pretrained models, we further increase adaptation difficulty by training on and evaluating adaptation performance on all tasks within a task suite simultaneously.\footnote{\iclrnew{We use one adapter per task \emph{suite}}. LIBERO~\citep{liu2023libero} originally evaluated on a per-task basis.}
We pretrain on \textbf{Kitchen} until performance convergence (100 epochs). Subsequent adaptations follow two setups: (1) sequential adaptation across the \textbf{Spatial}, \textbf{Goal}, \textbf{Object}, \textbf{Living Room}, and \textbf{Study Room} task suites for 100 epochs each, and (2) adaptation to each long-horizon task within the \textbf{LIBERO-10} benchmark over 50 epochs.
Each experiment is conducted with 3 different random seeds.
Except for the Experience Replay (ER) method, data from earlier tasks remains unavailable in later stages.
Our diverse adaptation setup provides a thorough and in-depth examination of knowledge transfer across a spectrum of domains, including spatial, procedural, visual, and compositional.

In the pretraining phase for \method, \iclrnew{we add trainable adapters to the CLIP spatial and instruction encoders while freezing the encoder weights}. \iclrnew{All other model weights are fully learnable.} 
During adaptation, \iclrnew{the CLIP encoders and the GPT2 decoder are frozen, while adapters for them, the fusion module, and the policy head are tuned}. Adapter weights are initialized from previous adapters with minor random noise. A fusion module and policy head copy are maintained during the adaptation for both \method\ and FPF.
The detailed hyperparameters are presented in Appendix \ref{sec:appendix:models}.

\vspace{-3mm}
\subsection{Results and analysis}
\label{sec:result}
\vspace{-2mm}

\textbf{Comparison of \method\ Integration Styles.}
Fig. \ref{fig:training_curve_adapter} showcases the continual adaptation success rates for different \method\ methods. 
The efficacy of LoRA suggests that a well-pretrained model has a surprisingly low intrinsic dimension for imitation learning tasks~\citep{aghajanyan2020intrinsic}. This implies the existence of a low-rank reparameterization that is just as adept for fine-tuning as the full parameter space. Further, the prefix tuning method outperforms the bottleneck-based approach~\citep{houlsby2019parameter}, indicating that the sequential integration style may not be the optimal choice for continual learning, potentially due to its inherent "filtering" mechanism. Surprisingly, RoboAdapter~\citep{sharma2023lossless} generally performs the worst, potentially due to only introducing weights after the feedforward layer as opposed to after~\citep{houlsby2019parameter} or within~\citep{li2021prefix, hu2021lora} the attention layer. 
Due to LoRA's pronounced effectiveness, it is predominantly employed as our \method\ integration method in subsequent experiments.

\textbf{\method\ vs. Conventional Fine-tuning.}
Across all evaluations, \method\ vastly outperforms all baselines in both forward and backward transfer, demonstrating that conventional fine-tuning methods are weak in data-scarce continual learning.
In Fig.~\ref{fig:training_curve_main} we plot continual learning success rates over 6 task suites, where \method\ outperforms the best baselines by over \textbf{3x} in some comparisons and generally achieves the best success rates.
We display additional results on LIBERO-10, long-horizon tasks, in Table~\ref{tab:long-tasks-results}. Here, \method\ again performs best, with perfect backward transfer and forward transfer capabilities significantly better than the baselines:
FFT not only exhibits marked catastrophic forgetting of earlier tasks---evidenced by poor BWT---but also compromises the model's adaptability to new tasks. 
This decline in forward transfer is characterized by a steady descent in success rates as training progresses, displayed in Appendix Table~\ref{tab:long-tasks-results-appendix}.
Such deterioration in flexibility has been recognized in other studies as well~\citep{lyle2022understanding, kumar2023maintaining}.
\iclrnew{PackNet is able to adapt well on some task suites as it learns new parameters within different parts of the model, but overall is still outperformed by \method.}

\begin{figure}[t]
    \centering
    \includegraphics[width=\textwidth]{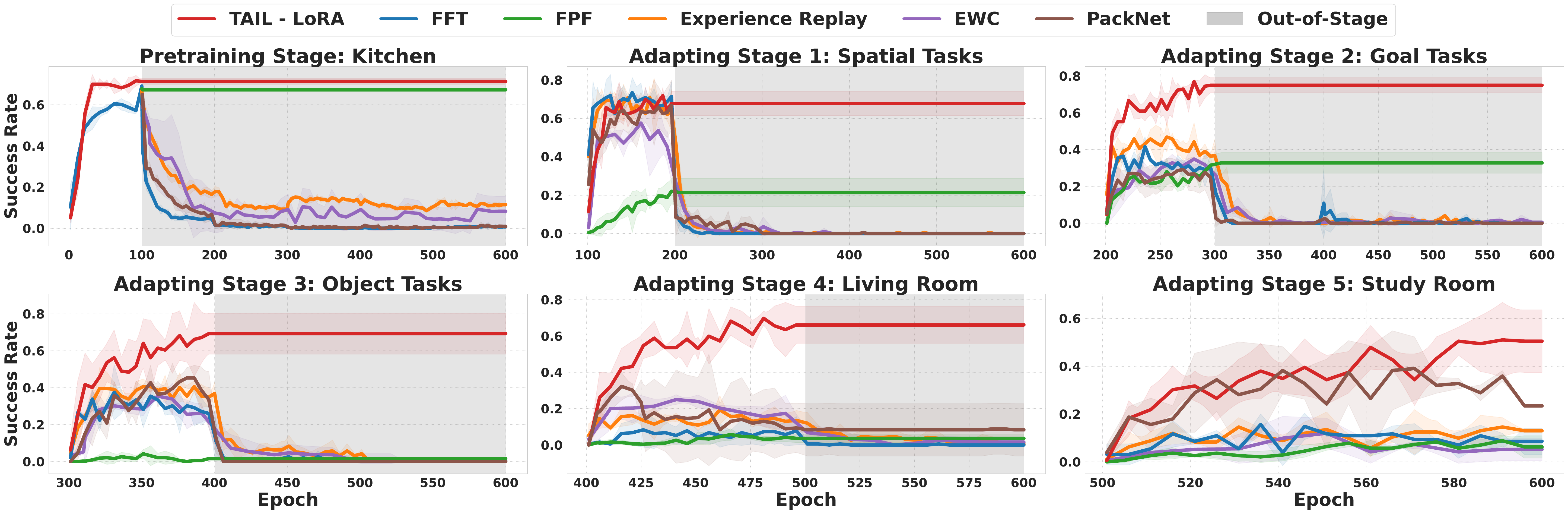}
    \vspace{-4mm}
        \caption{\small Success rates on the pretraining stage on 40 tasks in the LIBERO Kitchen scene and 5 adaptation stages, each with 8 tasks over 100 epochs, which are continuously evaluated in subsequent stages (\textcolor{gray!90}{shaded area}). }%
        \vspace{-5mm}
    \label{fig:training_curve_main}
\end{figure}

\begin{table}[!h]
    \centering
    \caption{\small Adaptation results on 10 long horizon tasks, higher is better. The BWT for TAIL methods are all 0 (no forgetting). FPF results were omitted due to its near-zero performance. See per-task results in Appendix Table~\ref{tab:long-tasks-results-appendix}.}
    \label{tab:long-tasks-results}
    \small
    \renewcommand{\arraystretch}{1.1}
    \resizebox{\linewidth}{!}{
        \begin{tabular}{@{}lcccccc>{\columncolor[gray]{0.9}}c>{\columncolor[gray]{0.9}}c>{\columncolor[gray]{0.9}}c>{\columncolor[gray]{0.9}}c>{\columncolor[gray]{0.9}}c@{}}
            \toprule
            \multirow{3}{*}{} & \multicolumn{6}{c}{Conventional Fine-Tuning Methods} & \multicolumn{4}{c}{TAIL-based Methods (\textbf{Ours})} \\ \cmidrule(lr){2-7} \cmidrule(lr){8-11}
            & \multicolumn{2}{c}{Full Fine-Tuning} & \multicolumn{2}{c}{Experience Replay} & \multicolumn{2}{c}{EWC} & \multicolumn{1}{c}{LoRA} & \multicolumn{1}{c}{Prefix} & \multicolumn{1}{c}{Bottleneck} & \multicolumn{1}{c}{RoboAdapter} \\
            & FWT $\uparrow$ & BWT $\uparrow$ & FWT $\uparrow$ & BWT $\uparrow$ & FWT $\uparrow$ & BWT $\uparrow$ & \multicolumn{1}{c}{FWT $\uparrow$} & \multicolumn{1}{c}{FWT $\uparrow$} & \multicolumn{1}{c}{FWT $\uparrow$} & \multicolumn{1}{c}{FWT $\uparrow$} \\ \midrule
            Average & 0.48 \scriptsize$\pm$ 0.10 & -0.55 \scriptsize$\pm$ 0.21 & 0.45 \scriptsize$\pm$ 0.09 & -0.49 \scriptsize$\pm$ 0.23 & 0.30 \scriptsize$\pm$ 0.16 & -0.43 \scriptsize$\pm$ 0.20 & \textbf{0.70 \scriptsize$\pm$ 0.10} & 0.51 \scriptsize$\pm$ 0.15 & 0.46 \scriptsize$\pm$ 0.11 & 0.42 \scriptsize$\pm$ 0.13 \\ \bottomrule
        \end{tabular}
    }
\end{table}

\vspace{-3mm}
\textbf{\iclrnew{Adaptation Plasticity.}} Exhaustive fine-tuning on specialized domains has been found to distort pretrained features~\citep{kumar2022finetuning}, undermining model adaptability. Our circle-back experiments in Table \ref{tab:circle back}, where a full fine-tuned model is re-trained on prior task suites, demonstrate a steep performance drop upon re-visiting previously learned tasks. \iclrnew{Additional experiments in Appendix~\ref{sec:appendix:more_results:different_base_models} further highlight this issue.}

\begin{wraptable}[9]{R}{6cm}
\vspace{-0.4cm}
\centering
\caption{\small The success rate of initial training and revisiting previous tasks with FFT. FFT suffers from catastrophic forgetting and performs worse on re-visits despite re-training on the same data.
}
\label{tab:circle back}
\large
\resizebox{1.\linewidth}{!}{
\begin{tabular}{@{}lccc@{}}
\toprule \multirow{2}{*}{Type} & \multicolumn{3}{c}{LIBERO Task Suite} \\ 
            \cmidrule(lr){2-4}
  & Spatial                               & Goal                                  & Object                                \\ \hline
Initial  & 0.79                                  & 0.42                                  & 0.42                                  \\ \hline
Re-visit & 0.53 \small \textcolor[HTML]{FF0000}{( $\downarrow$ 0.26)} & 0.20 \small\textcolor[HTML]{FF0000}{($\downarrow$0.22)} & 0.27 \small\textcolor[HTML]{FF0000}{($\downarrow$0.15)} \\ \hline
\end{tabular}
}
\end{wraptable}

The training and validation losses, detailed in Appendix \ref{sec:appendix:more_results:overfitting} and Fig.~\ref{fig:loss_curve}, highlight FFT's propensity to overfit. 
This translates to a notable decline in success rates, reinforcing the challenges FFT faces in balancing retention of prior tasks with the assimilation of new ones.

While ER and the regularization-based method EWC exhibit some potential in mitigating catastrophic forgetting, they were detrimental to forward transfer performance. 
Their downsides are also reflected in storage and computing costs: ER requires more storage for previous data than \method\ LoRA adapter weights (e.g., Kitchen dataset at 28GB vs 7.8MB for \method's LoRA adapter).
Furthermore, EWC presents significant challenges for larger models because of the increased GPU memory consumption from maintaining a copy of the entire weights of the old model in memory. %
We also found it to exhibit unstable training due to the regularization loss. More discussions are presented in Appendix \ref{sec:appendix:baselines}.

\textbf{When does \method\ work best?} The efficacy of \method\ hinges significantly on the base model's features. We compare \method\ under different pretraining strategies and models in Appendix Sec.~\ref{sec:appendix:more_results:pretrained_influence} and ~\ref{sec:appendix:more_results:different_base_models}. In short, \method\ works best with our pretraining architecture and frozen CLIP visual/language encoders, and performance drops when we fine-tune the pretrained encoders, likely as FFT contaminates the rich CLIP features when fine-tuned in a niche domain with sparse data. 

\begin{table}[]
\centering
\caption{\small Comparison of trainable parameters and memory usage for \method\ and FFT. We use \textcolor[HTML]{008000}{($\cdot$\%)} and \textcolor[HTML]{EE82EE}{$\downarrow$ ($\cdot$\%)} to denote the percentage of trainable parameter and the decrease of GPU memory w.r.t FFT.
}
\label{tab:parameters}
\large
\resizebox{1.\linewidth}{!}{
\begin{tabular}{@{}lccccc@{}}
\toprule \multirow{2}{*}{Method} & Conventional & \multicolumn{4}{c}{TAIL-based Methods (Ours)} \\ 
            \cmidrule(lr){3-6}
            \cmidrule(lr){2-2}
          & Full Fine-Tuning & LoRA           & RoboAdapter    & Bottleneck Adapter & Prefix Tuning  \\ \hline
CLIP (Spatial \& Task Encoder)             & 149.62M    & 0.49M          & 1.29M          & 1.31M              & 0.58M          \\
GPT2 (Temporal Encoder)           & 21.78M     & 0.69M          & 0.40M          & 0.40M              & 0.24M          \\
Fusion module and policy head   & 0.84M      & 0.84M          & 0.84M & 0.84M              & 0.84M          \\ \hline
Total Parameters & 172.24M    & 2.02M \textcolor[HTML]{008000}{(1.17\%)} & 2.53M \textcolor[HTML]{008000}{(1.47\%)} & 2.55M \textcolor[HTML]{008000}{(1.48\%)}     & 1.66M \textcolor[HTML]{008000}{(0.93\%)} \\ 
GPU Memory (Batch 14) & 20.1G    & 15.5G \textcolor[HTML]{EE82EE}{($\downarrow$ 23\%)} & 14.0G \textcolor[HTML]{EE82EE}{($\downarrow$ 30\%)} & 14.9G  \textcolor[HTML]{EE82EE}{($\downarrow$ 26\%)}    & 15.8G \textcolor[HTML]{EE82EE}{($\downarrow$ 21\%)} \\ \hline
\end{tabular}
}
\end{table}

\textbf{Analysis Summary.} 
We argue in favor of a large pretrained base model augmented with numerous lightweight plugins tailored for different downstream tasks. This framework, \method, holds considerable promise for advancing embodied intelligence in real-world applications; the storage footprint of our entire model is about 660MB, and duplicating this model for each task in a stream of oncoming tasks is impractical. Meanwhile, the space occupied by one such model can accommodate as many as 84 task-specific adapters, which, as our experiments show, can outperform full fine-tuning regardless.
Moreover, the features of the pretrained weights remain intact, ensuring their applicability across a broad array of domains.
In summary, \method\ offers a promising avenue for the efficient adaptation of large decision-making models. Despite the fact that our method requires significantly less computation and memory (and storage), our experiments show that it consistently outperforms all prior approaches in the continual learning setting. We would also like to highlight that the TAIL framework is not restricted to imitation learning, but also other learning methods such as reinforcement learning.

\vspace{-2mm}
\section{Conclusion}
\vspace{-3mm}
In this study, we examined the challenges of efficiently adapting large pretrained models for decision-making and robotics applications. We proposed \method, an efficient adaptation framework for pretrained decision-making models. Through a comprehensive exploration of parameter-efficient fine-tuning (PEFT) techniques in \method, especially Low-Rank Adaptation (LoRA), we demonstrated their potential in enhancing adaptation efficiency, mitigating catastrophic forgetting, and ensuring robust performance across diverse tasks. Our empirical evaluations on the LIBERO benchmark further underscored the advantages of these techniques in continual learning scenarios. As the demand for adaptive, intelligent agents grows across various domains, the insights from this research offer a promising direction for the future of efficient model adaptation in decision-making contexts.

\clearpage
\bibliography{main}

\begin{thebibliography}{75}
\providecommand{\natexlab}[1]{#1}
\providecommand{\url}[1]{\texttt{#1}}
\expandafter\ifx\csname urlstyle\endcsname\relax
  \providecommand{\doi}[1]{doi: #1}\else
  \providecommand{\doi}{doi: \begingroup \urlstyle{rm}\Url}\fi

\bibitem[Agarwal et~al.(2022)Agarwal, Schwarzer, Castro, Courville, and
  Bellemare]{agarwal2022reincarnating}
Rishabh Agarwal, Max Schwarzer, Pablo~Samuel Castro, Aaron~C Courville, and
  Marc Bellemare.
\newblock Reincarnating reinforcement learning: Reusing prior computation to
  accelerate progress.
\newblock \emph{Advances in Neural Information Processing Systems},
  35:\penalty0 28955--28971, 2022.

\bibitem[Aghajanyan et~al.(2020)Aghajanyan, Zettlemoyer, and
  Gupta]{aghajanyan2020intrinsic}
Armen Aghajanyan, Luke Zettlemoyer, and Sonal Gupta.
\newblock Intrinsic dimensionality explains the effectiveness of language model
  fine-tuning.
\newblock \emph{arXiv preprint arXiv:2012.13255}, 2020.

\bibitem[Bishop(1994)]{bishop1994mixture}
Christopher~M Bishop.
\newblock Mixture density networks.
\newblock 1994.

\bibitem[Bousmalis et~al.(2023)Bousmalis, Vezzani, Rao, Devin, Lee, Bauza,
  Davchev, Zhou, Gupta, Raju, et~al.]{bousmalis2023robocat}
Konstantinos Bousmalis, Giulia Vezzani, Dushyant Rao, Coline Devin, Alex~X Lee,
  Maria Bauza, Todor Davchev, Yuxiang Zhou, Agrim Gupta, Akhil Raju, et~al.
\newblock Robocat: A self-improving foundation agent for robotic manipulation.
\newblock \emph{arXiv preprint arXiv:2306.11706}, 2023.

\bibitem[Bozinovski \& Fulgosi(1976)Bozinovski and
  Fulgosi]{bozinovski1976influence}
Stevo Bozinovski and Ante Fulgosi.
\newblock The influence of pattern similarity and transfer learning upon
  training of a base perceptron b2.
\newblock In \emph{Proceedings of Symposium Informatica}, volume~3, pp.\
  121--126, 1976.

\bibitem[Brohan et~al.(2022)Brohan, Brown, Carbajal, Chebotar, Dabis, Finn,
  Gopalakrishnan, Hausman, Herzog, Hsu, et~al.]{brohan2022rt}
Anthony Brohan, Noah Brown, Justice Carbajal, Yevgen Chebotar, Joseph Dabis,
  Chelsea Finn, Keerthana Gopalakrishnan, Karol Hausman, Alex Herzog, Jasmine
  Hsu, et~al.
\newblock Rt-1: Robotics transformer for real-world control at scale.
\newblock \emph{arXiv preprint arXiv:2212.06817}, 2022.

\bibitem[Brohan et~al.(2023)Brohan, Brown, Carbajal, Chebotar, Chen,
  Choromanski, Ding, Driess, Dubey, Finn, Florence, Fu, Arenas,
  et~al.]{rt22023arxiv}
Anthony Brohan, Noah Brown, Justice Carbajal, Yevgen Chebotar, Xi~Chen,
  Krzysztof Choromanski, Tianli Ding, Danny Driess, Avinava Dubey, Chelsea
  Finn, Pete Florence, Chuyuan Fu, Montse~Gonzalez Arenas, et~al.
\newblock Rt-2: Vision-language-action models transfer web knowledge to robotic
  control.
\newblock In \emph{arXiv preprint arXiv:2307.15818}, 2023.

\bibitem[Brown et~al.(2020)Brown, Mann, Ryder, Subbiah, Kaplan, Dhariwal,
  Neelakantan, Shyam, Sastry, Askell, Agarwal, Herbert-Voss, Krueger, Henighan,
  Child, Ramesh, et~al.]{brown2020language}
Tom~B. Brown, Benjamin Mann, Nick Ryder, Melanie Subbiah, Jared Kaplan,
  Prafulla Dhariwal, Arvind Neelakantan, Pranav Shyam, Girish Sastry, Amanda
  Askell, Sandhini Agarwal, Ariel Herbert-Voss, Gretchen Krueger, Tom Henighan,
  Rewon Child, Aditya Ramesh, et~al.
\newblock Language models are few-shot learners, 2020.

\bibitem[Caccia et~al.(2023)Caccia, Mueller, Kim, Charlin, and
  Fakoor]{caccia2023taskagnostic}
Massimo Caccia, Jonas Mueller, Taesup Kim, Laurent Charlin, and Rasool Fakoor.
\newblock Task-agnostic continual reinforcement learning: Gaining insights and
  overcoming challenges.
\newblock In \emph{Conference on Lifelong Learning Agents}, 2023.

\bibitem[Chaudhry et~al.(2019)Chaudhry, Rohrbach, Elhoseiny, Ajanthan, Dokania,
  Torr, and Ranzato]{chaudhry2019tiny}
Arslan Chaudhry, Marcus Rohrbach, Mohamed Elhoseiny, Thalaiyasingam Ajanthan,
  Puneet~K Dokania, Philip~HS Torr, and Marc'Aurelio Ranzato.
\newblock On tiny episodic memories in continual learning.
\newblock \emph{arXiv preprint arXiv:1902.10486}, 2019.

\bibitem[Chebotar et~al.(2021)Chebotar, Hausman, Lu, Xiao, Kalashnikov, Varley,
  Irpan, Eysenbach, Julian, Finn, and Levine]{actionablemodels2021arxiv}
Yevgen Chebotar, Karol Hausman, Yao Lu, Ted Xiao, Dmitry Kalashnikov, Jake
  Varley, Alex Irpan, Benjamin Eysenbach, Ryan Julian, Chelsea Finn, and Sergey
  Levine.
\newblock Actionable models: Unsupervised offline reinforcement learning of
  robotic skills.
\newblock \emph{arXiv preprint arXiv:2104.07749}, 2021.

\bibitem[Chen et~al.(2021{\natexlab{a}})Chen, Liu, Zhu, Xu, Ding, Li, and
  Zhao]{chen2021context}
Baiming Chen, Zuxin Liu, Jiacheng Zhu, Mengdi Xu, Wenhao Ding, Liang Li, and
  Ding Zhao.
\newblock Context-aware safe reinforcement learning for non-stationary
  environments.
\newblock In \emph{2021 IEEE International Conference on Robotics and
  Automation (ICRA)}, pp.\  10689--10695. IEEE, 2021{\natexlab{a}}.

\bibitem[Chen et~al.(2021{\natexlab{b}})Chen, Lu, Rajeswaran, Lee, Grover,
  Laskin, Abbeel, Srinivas, and Mordatch]{chen2021decision}
Lili Chen, Kevin Lu, Aravind Rajeswaran, Kimin Lee, Aditya Grover, Misha
  Laskin, Pieter Abbeel, Aravind Srinivas, and Igor Mordatch.
\newblock Decision transformer: Reinforcement learning via sequence modeling.
\newblock \emph{Advances in neural information processing systems},
  34:\penalty0 15084--15097, 2021{\natexlab{b}}.

\bibitem[Chen et~al.(2022)Chen, Ge, Tong, Wang, Song, Wang, and
  Luo]{chen2022adaptformer}
Shoufa Chen, Chongjian Ge, Zhan Tong, Jiangliu Wang, Yibing Song, Jue Wang, and
  Ping Luo.
\newblock Adaptformer: Adapting vision transformers for scalable visual
  recognition.
\newblock \emph{Advances in Neural Information Processing Systems},
  35:\penalty0 16664--16678, 2022.

\bibitem[Collaboration et~al.(2023)Collaboration, Padalkar, Pooley, Jain,
  Bewley, Herzog, Irpan, Khazatsky, Rai, Singh, Brohan, Raffin, Wahid,
  et~al.]{open_x_embodiment_rt_x_2023}
Open X-Embodiment Collaboration, Abhishek Padalkar, Acorn Pooley, Ajinkya Jain,
  Alex Bewley, Alex Herzog, Alex Irpan, Alexander Khazatsky, Anant Rai, Anikait
  Singh, Anthony Brohan, Antonin Raffin, Ayzaan Wahid, et~al.
\newblock Open {X-E}mbodiment: Robotic learning datasets and {RT-X} models.
\newblock \url{https://robotics-transformer-x.github.io}, 2023.

\bibitem[Dietterich et~al.(1997)Dietterich, Pratt, and
  Thrun]{dietterich1997special}
Thomas~G Dietterich, Lorien Pratt, and Sebastian Thrun.
\newblock Special issue on inductive transfer.
\newblock \emph{Machine Learning}, 28\penalty0 (1), 1997.

\bibitem[Driess et~al.(2023)Driess, Xia, Sajjadi, Lynch, Chowdhery, Ichter,
  Wahid, Tompson, Vuong, Yu, Huang, et~al.]{driess2023palme}
Danny Driess, Fei Xia, Mehdi S.~M. Sajjadi, Corey Lynch, Aakanksha Chowdhery,
  Brian Ichter, Ayzaan Wahid, Jonathan Tompson, Quan Vuong, Tianhe Yu, Wenlong
  Huang, et~al.
\newblock Palm-e: An embodied multimodal language model.
\newblock In \emph{arXiv preprint arXiv:2303.03378}, 2023.

\bibitem[Fakoor et~al.(2020)Fakoor, Chaudhari, Soatto, and
  Smola]{fakoor2019meta}
Rasool Fakoor, Pratik Chaudhari, Stefano Soatto, and Alexander~J. Smola.
\newblock Meta-q-learning.
\newblock In \emph{International Conference on Learning Representations}, 2020.

\bibitem[Fu et~al.(2022)Fu, Yu, Littman, and Konidaris]{fu2022model}
Haotian Fu, Shangqun Yu, Michael Littman, and George Konidaris.
\newblock Model-based lifelong reinforcement learning with bayesian
  exploration.
\newblock \emph{Advances in Neural Information Processing Systems},
  35:\penalty0 32369--32382, 2022.

\bibitem[Grauman et~al.(2022)Grauman, Westbury, Byrne, Chavis, Furnari,
  Girdhar, Hamburger, Jiang, Liu, Liu, et~al.]{grauman2022ego4d}
Kristen Grauman, Andrew Westbury, Eugene Byrne, Zachary Chavis, Antonino
  Furnari, Rohit Girdhar, Jackson Hamburger, Hao Jiang, Miao Liu, Xingyu Liu,
  et~al.
\newblock Ego4d: Around the world in 3,000 hours of egocentric video.
\newblock In \emph{Proceedings of the IEEE/CVF Conference on Computer Vision
  and Pattern Recognition}, pp.\  18995--19012, 2022.

\bibitem[Gupta et~al.(2022)Gupta, Lynch, Kinman, Peake, Levine, and
  Hausman]{gupta2022dbap}
Abhishek Gupta, Corey Lynch, Brandon Kinman, Garrett Peake, Sergey Levine, and
  Karol Hausman.
\newblock Demonstration-bootstrapped autonomous practicing via multi-task
  reinforcement learning.
\newblock \emph{arXiv}, 2022.

\bibitem[Hansen et~al.(2022)Hansen, Yuan, Ze, Mu, Rajeswaran, Su, Xu, and
  Wang]{hansen2022on}
Nicklas Hansen, Zhecheng Yuan, Yanjie Ze, Tongzhou Mu, Aravind Rajeswaran, Hao
  Su, Huazhe Xu, and Xiaolong Wang.
\newblock On pre-training for visuo-motor control: Revisiting a
  learning-from-scratch baseline.
\newblock \emph{arXiv preprint arXiv:2212.05749}, 2022.

\bibitem[He et~al.(2022)He, Zhou, Ma, Berg-Kirkpatrick, and
  Neubig]{he2022towards}
Junxian He, Chunting Zhou, Xuezhe Ma, Taylor Berg-Kirkpatrick, and Graham
  Neubig.
\newblock Towards a unified view of parameter-efficient transfer learning.
\newblock In \emph{International Conference on Learning Representations}, 2022.

\bibitem[Houlsby et~al.(2019)Houlsby, Giurgiu, Jastrzebski, Morrone,
  De~Laroussilhe, Gesmundo, Attariyan, and Gelly]{houlsby2019parameter}
Neil Houlsby, Andrei Giurgiu, Stanislaw Jastrzebski, Bruna Morrone, Quentin
  De~Laroussilhe, Andrea Gesmundo, Mona Attariyan, and Sylvain Gelly.
\newblock Parameter-efficient transfer learning for nlp.
\newblock In \emph{International Conference on Machine Learning}, pp.\
  2790--2799. PMLR, 2019.

\bibitem[Hu et~al.(2021)Hu, Shen, Wallis, Allen-Zhu, Li, Wang, Wang, and
  Chen]{hu2021lora}
Edward~J Hu, Yelong Shen, Phillip Wallis, Zeyuan Allen-Zhu, Yuanzhi Li, Shean
  Wang, Lu~Wang, and Weizhu Chen.
\newblock Lora: Low-rank adaptation of large language models.
\newblock \emph{arXiv preprint arXiv:2106.09685}, 2021.

\bibitem[Jang et~al.(2021)Jang, Irpan, Khansari, Kappler, Ebert, Lynch, Levine,
  and Finn]{jang2021bcz}
Eric Jang, Alex Irpan, Mohi Khansari, Daniel Kappler, Frederik Ebert, Corey
  Lynch, Sergey Levine, and Chelsea Finn.
\newblock {BC}-z: Zero-shot task generalization with robotic imitation
  learning.
\newblock In \emph{5th Annual Conference on Robot Learning}, 2021.

\bibitem[Jiang et~al.(2023)Jiang, Gupta, Zhang, Wang, Dou, Chen, Fei-Fei,
  Anandkumar, Zhu, and Fan]{jiang2023vima}
Yunfan Jiang, Agrim Gupta, Zichen Zhang, Guanzhi Wang, Yongqiang Dou, Yanjun
  Chen, Li~Fei-Fei, Anima Anandkumar, Yuke Zhu, and Linxi Fan.
\newblock Vima: General robot manipulation with multimodal prompts.
\newblock In \emph{Fortieth International Conference on Machine Learning},
  2023.

\bibitem[Kalashnkov et~al.(2021)Kalashnkov, Varley, Chebotar, Swanson,
  Jonschkowski, Finn, Levine, and Hausman]{mtopt2021arxiv}
Dmitry Kalashnkov, Jake Varley, Yevgen Chebotar, Ben Swanson, Rico
  Jonschkowski, Chelsea Finn, Sergey Levine, and Karol Hausman.
\newblock Mt-opt: Continuous multi-task robotic reinforcement learning at
  scale.
\newblock \emph{arXiv}, 2021.

\bibitem[Ke et~al.(2020)Ke, Choudhury, Barnes, Sun, Lee, and
  Srinivasa]{ke2020imitation}
Liyiming Ke, Sanjiban Choudhury, Matt Barnes, Wen Sun, Gilwoo Lee, and
  Siddhartha Srinivasa.
\newblock Imitation learning as $f$-divergence minimization.
\newblock \emph{arXiv preprint 1905.12888}, 2020.

\bibitem[Kirkpatrick et~al.(2017)Kirkpatrick, Pascanu, Rabinowitz, Veness,
  Desjardins, Rusu, Milan, Quan, Ramalho, Grabska-Barwinska,
  et~al.]{kirkpatrick2017overcoming}
James Kirkpatrick, Razvan Pascanu, Neil Rabinowitz, Joel Veness, Guillaume
  Desjardins, Andrei~A Rusu, Kieran Milan, John Quan, Tiago Ramalho, Agnieszka
  Grabska-Barwinska, et~al.
\newblock Overcoming catastrophic forgetting in neural networks.
\newblock \emph{Proceedings of the national academy of sciences}, 114\penalty0
  (13):\penalty0 3521--3526, 2017.

\bibitem[Kumar et~al.(2022)Kumar, Raghunathan, Jones, Ma, and
  Liang]{kumar2022finetuning}
Ananya Kumar, Aditi Raghunathan, Robbie~Matthew Jones, Tengyu Ma, and Percy
  Liang.
\newblock Fine-tuning can distort pretrained features and underperform
  out-of-distribution.
\newblock In \emph{International Conference on Learning Representations}, 2022.

\bibitem[Kumar et~al.(2023)Kumar, Marklund, and Van~Roy]{kumar2023maintaining}
Saurabh Kumar, Henrik Marklund, and Benjamin Van~Roy.
\newblock Maintaining plasticity via regenerative regularization.
\newblock \emph{arXiv preprint arXiv:2308.11958}, 2023.

\bibitem[Lee et~al.(2023)Lee, Chen, Tajwar, Kumar, Yao, Liang, and
  Finn]{lee2022surgical}
Yoonho Lee, Annie~S Chen, Fahim Tajwar, Ananya Kumar, Huaxiu Yao, Percy Liang,
  and Chelsea Finn.
\newblock Surgical fine-tuning improves adaptation to distribution shifts.
\newblock \emph{International Conference on Learning Representations}, 2023.

\bibitem[Lester et~al.(2021)Lester, Al-Rfou, and Constant]{lester2021power}
Brian Lester, Rami Al-Rfou, and Noah Constant.
\newblock The power of scale for parameter-efficient prompt tuning.
\newblock \emph{arXiv preprint arXiv:2104.08691}, 2021.

\bibitem[Li \& Liang(2021)Li and Liang]{li2021prefix}
Xiang~Lisa Li and Percy Liang.
\newblock Prefix-tuning: Optimizing continuous prompts for generation.
\newblock \emph{arXiv preprint arXiv:2101.00190}, 2021.

\bibitem[Liang et~al.(2022)Liang, Singh, Pertsch, and
  Thomason]{liang2022transformer}
Anthony Liang, Ishika Singh, Karl Pertsch, and Jesse Thomason.
\newblock Transformer adapters for robot learning.
\newblock In \emph{CoRL 2022 Workshop on Pre-training Robot Learning}, 2022.

\bibitem[Liu et~al.(2023{\natexlab{a}})Liu, Zhu, Gao, Feng, Liu, Zhu, and
  Stone]{liu2023libero}
Bo~Liu, Yifeng Zhu, Chongkai Gao, Yihao Feng, Qiang Liu, Yuke Zhu, and Peter
  Stone.
\newblock Libero: Benchmarking knowledge transfer for lifelong robot learning.
\newblock \emph{arXiv preprint arXiv:2306.03310}, 2023{\natexlab{a}}.

\bibitem[Liu et~al.(2023{\natexlab{b}})Liu, Zheng, Du, Ding, Qian, Yang, and
  Tang]{liu2023gpt}
Xiao Liu, Yanan Zheng, Zhengxiao Du, Ming Ding, Yujie Qian, Zhilin Yang, and
  Jie Tang.
\newblock Gpt understands, too.
\newblock \emph{AI Open}, 2023{\natexlab{b}}.

\bibitem[Liu et~al.(2023{\natexlab{c}})Liu, Guo, Yao, Cen, Yu, Zhang, and
  Zhao]{liu2023constrained}
Zuxin Liu, Zijian Guo, Yihang Yao, Zhepeng Cen, Wenhao Yu, Tingnan Zhang, and
  Ding Zhao.
\newblock Constrained decision transformer for offline safe reinforcement
  learning.
\newblock \emph{arXiv preprint arXiv:2302.07351}, 2023{\natexlab{c}}.

\bibitem[Lopez-Paz \& Ranzato(2017)Lopez-Paz and Ranzato]{lopez2017gradient}
David Lopez-Paz and Marc'Aurelio Ranzato.
\newblock Gradient episodic memory for continual learning.
\newblock In \emph{Advances in Neural Information Processing Systems (NIPS)},
  2017.

\bibitem[Lu et~al.(2021)Lu, Hausman, Chebotar, Yan, Jang, Herzog, Xiao, Irpan,
  Khansari, Kalashnikov, and Levine]{awopt2021corl}
Yao Lu, Karol Hausman, Yevgen Chebotar, Mengyuan Yan, Eric Jang, Alexander
  Herzog, Ted Xiao, Alex Irpan, Mohi Khansari, Dmitry Kalashnikov, and Sergey
  Levine.
\newblock Aw-opt: Learning robotic skills with imitation andreinforcement at
  scale.
\newblock In \emph{5th Annual Conference on Robot Learning}, 2021.

\bibitem[Lyle et~al.(2022)Lyle, Rowland, and Dabney]{lyle2022understanding}
Clare Lyle, Mark Rowland, and Will Dabney.
\newblock Understanding and preventing capacity loss in reinforcement learning.
\newblock In \emph{International Conference on Learning Representations}, 2022.

\bibitem[Ma et~al.(2022)Ma, Sodhani, Jayaraman, Bastani, Kumar, and
  Zhang]{ma2022vip}
Yecheng~Jason Ma, Shagun Sodhani, Dinesh Jayaraman, Osbert Bastani, Vikash
  Kumar, and Amy Zhang.
\newblock Vip: Towards universal visual reward and representation via
  value-implicit pre-training.
\newblock \emph{arXiv preprint arXiv:2210.00030}, 2022.

\bibitem[Ma et~al.(2023)Ma, Liang, Som, Kumar, Zhang, Bastani, and
  Jayaraman]{ma2023liv}
Yecheng~Jason Ma, William Liang, Vaidehi Som, Vikash Kumar, Amy Zhang, Osbert
  Bastani, and Dinesh Jayaraman.
\newblock Liv: Language-image representations and rewards for robotic control.
\newblock \emph{arXiv preprint arXiv:2306.00958}, 2023.

\bibitem[Majumdar et~al.(2023{\natexlab{a}})Majumdar, Yadav, Arnaud, Ma, Chen,
  Silwal, Jain, Berges, Abbeel, Malik, Batra, Lin, Maksymets, Rajeswaran, and
  Meier]{vc2023}
Arjun Majumdar, Karmesh Yadav, Sergio Arnaud, Yecheng~Jason Ma, Claire Chen,
  Sneha Silwal, Aryan Jain, Vincent-Pierre Berges, Pieter Abbeel, Jitendra
  Malik, Dhruv Batra, Yixin Lin, Oleksandr Maksymets, Aravind Rajeswaran, and
  Franziska Meier.
\newblock Where are we in the search for an artificial visual cortex for
  embodied intelligence?
\newblock 2023{\natexlab{a}}.

\bibitem[Majumdar et~al.(2023{\natexlab{b}})Majumdar, Yadav, Arnaud, Ma, Chen,
  Silwal, Jain, Berges, Abbeel, Malik, et~al.]{majumdar2023we}
Arjun Majumdar, Karmesh Yadav, Sergio Arnaud, Yecheng~Jason Ma, Claire Chen,
  Sneha Silwal, Aryan Jain, Vincent-Pierre Berges, Pieter Abbeel, Jitendra
  Malik, et~al.
\newblock Where are we in the search for an artificial visual cortex for
  embodied intelligence?
\newblock \emph{arXiv preprint arXiv:2303.18240}, 2023{\natexlab{b}}.

\bibitem[Mallya \& Lazebnik(2018)Mallya and Lazebnik]{packnet2018}
Arun Mallya and Svetlana Lazebnik.
\newblock Packnet: Adding multiple tasks to a single network by iterative
  pruning.
\newblock In \emph{Proceedings of the IEEE Conference on Computer Vision and
  Pattern Recognition (CVPR)}, June 2018.

\bibitem[Mandlekar et~al.(2021)Mandlekar, Xu, Wong, Nasiriany, Wang, Kulkarni,
  Fei-Fei, Savarese, Zhu, and Mart{\'\i}n-Mart{\'\i}n]{mandlekar2021matters}
Ajay Mandlekar, Danfei Xu, Josiah Wong, Soroush Nasiriany, Chen Wang, Rohun
  Kulkarni, Li~Fei-Fei, Silvio Savarese, Yuke Zhu, and Roberto
  Mart{\'\i}n-Mart{\'\i}n.
\newblock What matters in learning from offline human demonstrations for robot
  manipulation.
\newblock \emph{arXiv preprint arXiv:2108.03298}, 2021.

\bibitem[Mao et~al.(2022)Mao, Mathias, Hou, Almahairi, Ma, Han, Yih, and
  Khabsa]{mao-etal-2022-unipelt}
Yuning Mao, Lambert Mathias, Rui Hou, Amjad Almahairi, Hao Ma, Jiawei Han,
  Wen-tau Yih, and Madian Khabsa.
\newblock Unipelt: A unified framework for parameter-efficient language model
  tuning.
\newblock 2022.

\bibitem[McCloskey \& Cohen(1989)McCloskey and
  Cohen]{McCloskey1989CatastrophicII}
Michael McCloskey and Neal~J Cohen.
\newblock Catastrophic interference in connectionist networks: The sequential
  learning problem.
\newblock In \emph{Psychology of learning and motivation}, volume~24, pp.\
  109--165. Elsevier, 1989.

\bibitem[Nair et~al.(2022)Nair, Rajeswaran, Kumar, Finn, and
  Gupta]{nair2022r3m}
Suraj Nair, Aravind Rajeswaran, Vikash Kumar, Chelsea Finn, and Abhinav Gupta.
\newblock R3m: A universal visual representation for robot manipulation.
\newblock \emph{arXiv preprint arXiv:2203.12601}, 2022.

\bibitem[Perez et~al.(2017)Perez, Strub, de~Vries, Dumoulin, and
  Courville]{perez2017film}
Ethan Perez, Florian Strub, Harm de~Vries, Vincent Dumoulin, and Aaron
  Courville.
\newblock Film: Visual reasoning with a general conditioning layer, 2017.

\bibitem[Pfeiffer et~al.(2020{\natexlab{a}})Pfeiffer, Kamath, R{\"u}ckl{\'e},
  Cho, and Gurevych]{pfeiffer2020adapterfusion}
Jonas Pfeiffer, Aishwarya Kamath, Andreas R{\"u}ckl{\'e}, Kyunghyun Cho, and
  Iryna Gurevych.
\newblock Adapterfusion: Non-destructive task composition for transfer
  learning.
\newblock \emph{arXiv preprint arXiv:2005.00247}, 2020{\natexlab{a}}.

\bibitem[Pfeiffer et~al.(2020{\natexlab{b}})Pfeiffer, R{\"u}ckl{\'e}, Poth,
  Kamath, Vuli{\'c}, Ruder, Cho, and Gurevych]{pfeiffer2020adapterhub}
Jonas Pfeiffer, Andreas R{\"u}ckl{\'e}, Clifton Poth, Aishwarya Kamath, Ivan
  Vuli{\'c}, Sebastian Ruder, Kyunghyun Cho, and Iryna Gurevych.
\newblock Adapterhub: A framework for adapting transformers.
\newblock \emph{arXiv preprint arXiv:2007.07779}, 2020{\natexlab{b}}.

\bibitem[Radford et~al.(2019{\natexlab{a}})Radford, Wu, Child, Luan, Amodei,
  and Sutskever]{radford2019GPT2}
Alec Radford, Jeff Wu, Rewon Child, David Luan, Dario Amodei, and Ilya
  Sutskever.
\newblock Language models are unsupervised multitask learners.
\newblock 2019{\natexlab{a}}.

\bibitem[Radford et~al.(2019{\natexlab{b}})Radford, Wu, Child, Luan, Amodei,
  Sutskever, et~al.]{radford2019language}
Alec Radford, Jeffrey Wu, Rewon Child, David Luan, Dario Amodei, Ilya
  Sutskever, et~al.
\newblock Language models are unsupervised multitask learners.
\newblock \emph{OpenAI blog}, 1\penalty0 (8):\penalty0 9, 2019{\natexlab{b}}.

\bibitem[Radford et~al.(2021)Radford, Kim, Hallacy, Ramesh, Goh, Agarwal,
  Sastry, Askell, Mishkin, Clark, et~al.]{radford2021learning}
Alec Radford, Jong~Wook Kim, Chris Hallacy, Aditya Ramesh, Gabriel Goh,
  Sandhini Agarwal, Girish Sastry, Amanda Askell, Pamela Mishkin, Jack Clark,
  et~al.
\newblock Learning transferable visual models from natural language
  supervision.
\newblock In \emph{International conference on machine learning}, pp.\
  8748--8763. PMLR, 2021.

\bibitem[Rebuffi et~al.(2018)Rebuffi, Vedaldi, and Bilen]{rebuffi2018}
S.~Rebuffi, A.~Vedaldi, and H.~Bilen.
\newblock Efficient parametrization of multi-domain deep neural networks.
\newblock In \emph{2018 IEEE/CVF Conference on Computer Vision and Pattern
  Recognition (CVPR)}, pp.\  8119--8127, Los Alamitos, CA, USA, jun 2018. IEEE
  Computer Society.
\newblock \doi{10.1109/CVPR.2018.00847}.
\newblock URL
  \url{https://doi.ieeecomputersociety.org/10.1109/CVPR.2018.00847}.

\bibitem[Reed et~al.(2022)Reed, Zolna, Parisotto, Colmenarejo, Novikov,
  Barth-maron, Gim{\'e}nez, Sulsky, et~al.]{reed2022gato}
Scott Reed, Konrad Zolna, Emilio Parisotto, Sergio~G{\'o}mez Colmenarejo,
  Alexander Novikov, Gabriel Barth-maron, Mai Gim{\'e}nez, Yury Sulsky, et~al.
\newblock A generalist agent.
\newblock \emph{Transactions on Machine Learning Research}, 2022.
\newblock ISSN 2835-8856.
\newblock Featured Certification, Outstanding Certification.

\bibitem[Rolnick et~al.(2019)Rolnick, Ahuja, Schwarz, Lillicrap, and
  Wayne]{rolnick2019experience}
David Rolnick, Arun Ahuja, Jonathan Schwarz, Timothy Lillicrap, and Gregory
  Wayne.
\newblock Experience replay for continual learning.
\newblock \emph{Advances in Neural Information Processing Systems}, 32, 2019.

\bibitem[Ross et~al.(2011)Ross, Gordon, and Bagnell]{ross2011dagger}
Stephane Ross, Geoffrey Gordon, and Drew Bagnell.
\newblock A reduction of imitation learning and structured prediction to
  no-regret online learning.
\newblock In \emph{Proceedings of the Fourteenth International Conference on
  Artificial Intelligence and Statistics}, volume~15 of \emph{Proceedings of
  Machine Learning Research}, pp.\  627--635. PMLR, 11--13 Apr 2011.

\bibitem[Schmidhuber(1992)]{schmidhuber1992learning}
Jürgen Schmidhuber.
\newblock Learning complex, extended sequences using the principle of history
  compression.
\newblock \emph{Neural Computation}, 4\penalty0 (2):\penalty0 234--242, 1992.
\newblock \doi{10.1162/neco.1992.4.2.234}.

\bibitem[Schmied et~al.(2023)Schmied, Hofmarcher, Paischer, Pascanu, and
  Hochreiter]{schmied2023learning}
Thomas Schmied, Markus Hofmarcher, Fabian Paischer, Razvan Pascanu, and Sepp
  Hochreiter.
\newblock Learning to modulate pre-trained models in rl.
\newblock \emph{arXiv preprint arXiv:2306.14884}, 2023.

\bibitem[Shafiullah et~al.(2022)Shafiullah, Cui, Altanzaya, and
  Pinto]{shafiullah2022behavior}
Nur Muhammad~Mahi Shafiullah, Zichen~Jeff Cui, Ariuntuya Altanzaya, and Lerrel
  Pinto.
\newblock Behavior transformers: Cloning \$k\$ modes with one stone.
\newblock In Alice~H. Oh, Alekh Agarwal, Danielle Belgrave, and Kyunghyun Cho
  (eds.), \emph{Advances in Neural Information Processing Systems}, 2022.

\bibitem[Sharma et~al.(2023)Sharma, Fantacci, Zhou, Koppula, Heess, Scholz, and
  Aytar]{sharma2023lossless}
Mohit Sharma, Claudio Fantacci, Yuxiang Zhou, Skanda Koppula, Nicolas Heess,
  Jon Scholz, and Yusuf Aytar.
\newblock Lossless adaptation of pretrained vision models for robotic
  manipulation.
\newblock \emph{arXiv preprint arXiv:2304.06600}, 2023.

\bibitem[Shridhar et~al.(2022)Shridhar, Manuelli, and Fox]{shridhar2022cliport}
Mohit Shridhar, Lucas Manuelli, and Dieter Fox.
\newblock Cliport: What and where pathways for robotic manipulation.
\newblock In \emph{Conference on Robot Learning}, pp.\  894--906. PMLR, 2022.

\bibitem[Thrun \& Mitchell(1995)Thrun and Mitchell]{Thrun95}
Sebastian Thrun and Tom~M Mitchell.
\newblock Lifelong robot learning.
\newblock In \emph{The biology and technology of intelligent autonomous
  agents}, pp.\  165--196. Springer, 1995.

\bibitem[Touvron et~al.(2023)Touvron, Lavril, Izacard, Martinet, Lachaux,
  Lacroix, Rozière, Goyal, Hambro, Azhar, Rodriguez, Joulin, Grave, and
  Lample]{touvron2023llama}
Hugo Touvron, Thibaut Lavril, Gautier Izacard, Xavier Martinet, Marie-Anne
  Lachaux, Timothée Lacroix, Baptiste Rozière, Naman Goyal, Eric Hambro,
  Faisal Azhar, Aurelien Rodriguez, Armand Joulin, Edouard Grave, and Guillaume
  Lample.
\newblock Llama: Open and efficient foundation language models, 2023.

\bibitem[Traor{\'{e}} et~al.(2019)Traor{\'{e}}, Caselles{-}Dupr{\'{e}}, Lesort,
  Sun, Cai, Rodr{\'{\i}}guez, and Filliat]{Traore19DisCoRL}
Ren{\'{e}} Traor{\'{e}}, Hugo Caselles{-}Dupr{\'{e}}, Timoth{\'{e}}e Lesort,
  Te~Sun, Guanghang Cai, Natalia~D{\'{\i}}az Rodr{\'{\i}}guez, and David
  Filliat.
\newblock Discorl: Continual reinforcement learning via policy distillation.
\newblock \emph{CoRR}, abs/1907.05855, 2019.

\bibitem[Vaswani et~al.(2017)Vaswani, Shazeer, Parmar, Uszkoreit, Jones, Gomez,
  Kaiser, and Polosukhin]{vaswani2017attention}
Ashish Vaswani, Noam Shazeer, Niki Parmar, Jakob Uszkoreit, Llion Jones,
  Aidan~N Gomez, \L~ukasz Kaiser, and Illia Polosukhin.
\newblock Attention is all you need.
\newblock In I.~Guyon, U.~Von Luxburg, S.~Bengio, H.~Wallach, R.~Fergus,
  S.~Vishwanathan, and R.~Garnett (eds.), \emph{Advances in Neural Information
  Processing Systems}, volume~30. Curran Associates, Inc., 2017.

\bibitem[Xu et~al.(2022)Xu, Shen, Zhang, Lu, Zhao, Tenenbaum, and
  Gan]{xu2022prompting}
Mengdi Xu, Yikang Shen, Shun Zhang, Yuchen Lu, Ding Zhao, Joshua Tenenbaum, and
  Chuang Gan.
\newblock Prompting decision transformer for few-shot policy generalization.
\newblock In \emph{international conference on machine learning}, pp.\
  24631--24645. PMLR, 2022.

\bibitem[Xu et~al.(2023)Xu, Lu, Shen, Zhang, Zhao, and Gan]{xu2023hyper}
Mengdi Xu, Yuchen Lu, Yikang Shen, Shun Zhang, Ding Zhao, and Chuang Gan.
\newblock Hyper-decision transformer for efficient online policy adaptation.
\newblock \emph{arXiv preprint arXiv:2304.08487}, 2023.

\bibitem[Yao et~al.(2024)Yao, Liu, Cen, Zhu, Yu, Zhang, and
  Zhao]{yao2024constraint}
Yihang Yao, Zuxin Liu, Zhepeng Cen, Jiacheng Zhu, Wenhao Yu, Tingnan Zhang, and
  Ding Zhao.
\newblock Constraint-conditioned policy optimization for versatile safe
  reinforcement learning.
\newblock \emph{Advances in Neural Information Processing Systems}, 36, 2024.

\bibitem[Zhang et~al.(2023{\natexlab{a}})Zhang, Pertsch, Zhang, and
  Lim]{zhang2023sprint}
Jesse Zhang, Karl Pertsch, Jiahui Zhang, and Joseph~J. Lim.
\newblock Sprint: Scalable policy pre-training via language instruction
  relabeling.
\newblock \emph{arXiv preprint arXiv:2306.11886}, 2023{\natexlab{a}}.

\bibitem[Zhang et~al.(2023{\natexlab{b}})Zhang, Zhang, Pertsch, Liu, Ren,
  Chang, Sun, and Lim]{zhang2023bootstrap}
Jesse Zhang, Jiahui Zhang, Karl Pertsch, Ziyi Liu, Xiang Ren, Minsuk Chang,
  Shao-Hua Sun, and Joseph~J Lim.
\newblock Bootstrap your own skills: Learning to solve new tasks with large
  language model guidance.
\newblock In \emph{7th Annual Conference on Robot Learning},
  2023{\natexlab{b}}.

\end{thebibliography}
\bibliographystyle{iclr2024_conference}
\clearpage
\appendix
\begin{center}
{\LARGE  Appendix: TAIL: Task-specific adapters for imitation learning with large pretrained models
}
\end{center}

\section{Model Architecture Details}
\label{sec:appendix:architecture}

\begin{figure}[h]
\centering
\includegraphics[width=\linewidth]{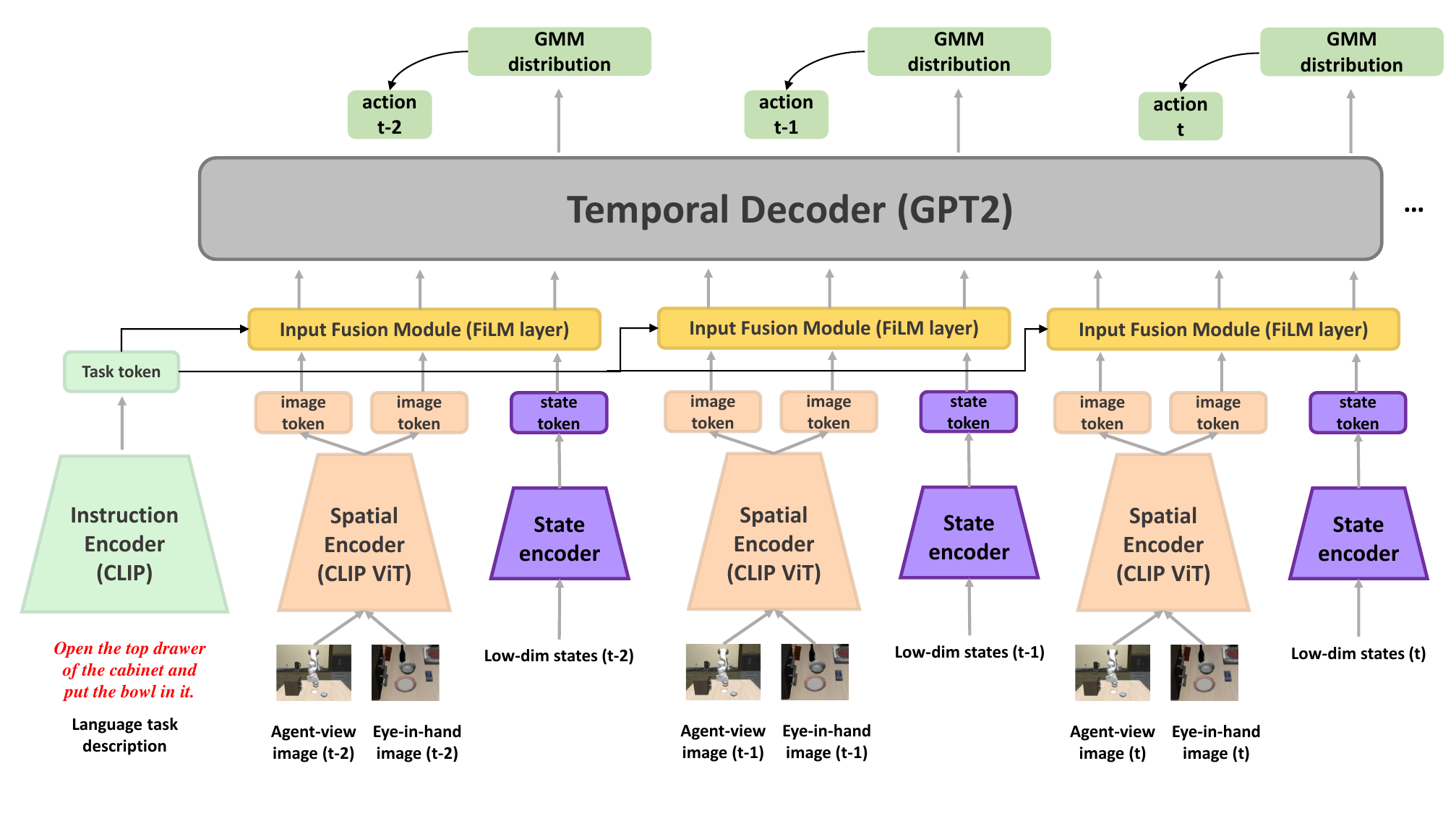}
\caption{A detailed view of the multi-modal, transformer policy architecture we utilize for pretraining. We encode language task descriptions with a pretrained CLIP \textcolor{green}{instruction encoder} and image observations with a pretrained CLIP \textcolor[HTML]{F7921D}{spatial encoder}. We additionally encode robot state observations which, along with the observation embeddings, are embedded into a sequence of tokens used by the \textcolor[HTML]{C8D7FF}{temporal decoder} transformer to predict single-step action distributions. We include an \textcolor[HTML]{FFD966}{input fusion module} (FiLM~\citep{perez2017film}) to explicitly combine the task embedding with the observation and state embeddings for better instruction-following ability.}
\end{figure}

\subsection{Pretrained Input Encoders}
We utilize pretrained CLIP image and textual encoders~\citep{radford2021learning} to encode image observations and language goal descriptions, respectively. Note that we do not use a pre-trained encoder for the low-dimensional state; the state encoder is learned from scratch.
\subsection{Input Modality Fusion}\label{appx:fusion}
We utilize Feature-wise Linear Modulation (FiLM) layers~\citep{perez2017film} (Fig.~\ref{fig:model-arch}(a), \textcolor[HTML]{FDBC42}{input fusion module}) to fuse language task specifications with image observations.
FiLM is a technique in multi-modal deep learning which modulates the intermediate activations of a neural network based on external information. Rather than explicitly designing architectures for conditional computation, FiLM layers simply use the features from one network to modulate the features of another. 

Let's consider a neural network $f$ with intermediate activations $x$ and an external network $g$ which outputs modulation parameters $\gamma$ and $\beta$. The modulated features $x'$ are given by:

\begin{align}
\gamma, \beta &= g(z) \\
x' &= \gamma \odot x + \beta,
\end{align}

where $z$ is the input to the external network $g$; $\odot$ represents element-wise multiplication; $\gamma$ and $\beta$ are vectors having the same size as $x$, with each element modulating a corresponding feature in $x$.

Thus, FiLM layers allow for a dynamic and feature-wise conditional computation without needing explicit architectural changes.
As such, task token (language) embeddings are given as input to a fully connected feedforward network, which outputs scale and translation parameters for the image and state embeddings.
These parameters modulate the image and state embeddings before they are passed to the transformer backbone.

\subsection{Temporal Transformer Backbone}
We utilize a standard GPT-2~\citep{radford2019GPT2} transformer backbone for our policy. 
Its input is a sequence of image and low-dim state encodings (robot joint states in LIBERO) and it outputs an action distribution.
Following the literature \citep{mandlekar2021matters, liu2023libero}, we adopt a stochastic policy parametrization based on a Gaussian-Mixture-Model (GMM)  \citep{bishop1994mixture}. 
Therefore, for every decision-making step, the transformer produces a latent vector of Gaussian means and variances, one for each of the GMM modes.
We optimize the parameters of the model with the negative log-likelihood loss on the ground truth actions based on the parameters of the GMM distribution.
At evaluation time, we deterministically select the next action parameterized by the mean of the Gaussian model with the highest density.

The environment configuration and the temporal decoder (GPT-2) hyperparameters are presented in Table \ref{tab:env config}.

\begin{table}[h]
\centering
\caption{\small Environment configuration and GPT-2 model hyperparameters}
\renewcommand{\arraystretch}{1.1}
\resizebox{1.\linewidth}{!}{
\begin{tabular}{ccccccccc}
\cline{1-2} \cline{4-7}
\multicolumn{2}{c}{\textbf{Environment Configuration}} & & \multicolumn{4}{c}{\textbf{GPT2 Temporal Encoder Configuration}} \\
\cline{1-2} \cline{4-7}
Action Dim. & 7 & & Max Seq Length & 8 & Activation & Gelu New \\
Raw State Dim. & 9 & & Number of Heads & 8 & Number of Layers & 6 \\
Max Episode Length & 500 & & GMM Min Std & 0.0001 & GMM Modes & 5 \\
Image Resolution & 128 x 128 & & FiLM Layers & 2 & Dropout & 0.15 \\
Image Views & Agent Front, Eye-in-Hand & & & & & \\
\cline{1-2} \cline{4-7}
\end{tabular}
}
\label{tab:env config}
\end{table}

\section{Implementation and Training Details}
\label{sec:appendix:models}

\subsection{Baseline Details}
\label{sec:appendix:baselines}

\textbf{Experience Replay (ER).} 
ER \citep{chaudhry2019tiny, rolnick2019experience} is a rehearsal-based approach that retains a buffer of samples from previous tasks to facilitate the learning of new tasks. After completing the learning process for a task, a subset of the data is saved into this buffer. During the training of subsequent tasks, ER draws samples from this buffer and mixes them with current task data. This process ensures that the training data closely resembles the distribution of data across all tasks. In our setup, we store all the previous trajectories in a replay buffer. For each training iteration on a new task, we uniformly sample $50\%$ trajectories from this buffer and $50\%$ from the new task's training data, respectively.

\textbf{Elastic Weight Consolidation (EWC).} EWC \citep{kirkpatrick2017overcoming} is a regularization method that adds a term to the standard single-task learning objective to constrain the updates of the neural network. This constraint uses the Fisher information matrix to gauge the significance of each network parameter. The loss function for task \( k \) is represented as:
\[ L_{\text{EWC}_k}(\theta) = L_{\text{BC}_K}(\theta) + \sum_{i} \frac{\lambda}{2} F_i (\theta_i - \theta_{k-1,i}^*)^2 \]
Here, \( \lambda \) is a hyperparameter penalty, and \( F_i \) is the diagonal of the Fisher information matrix given by:
\[ F_k = \mathbb{E}_{s \sim D_k, a \sim p_\theta(\cdot|s)} \left( \nabla_{\theta_k} \log p_{\theta_k}(a|s) \right)^2 \]
For our experiments, we adopt the online version of EWC. It updates the Fisher information matrix using an exponential moving average throughout the lifelong learning process. The actual Fisher Information Matrix estimate used is:
\[ \tilde{F}_k = \gamma F_{k-1} + (1 - \gamma)F_k \]
with \( F_k = \mathbb{E}_{(s,a) \sim D_k} \left( \nabla_{\theta_k} \log p_{\theta_k}(a|s) \right)^2 \) and \( k \) representing the task number. Following the benchmark implementation \citep{liu2023libero}, the hyperparameters are set as \( \gamma = 0.9 \) and \( \lambda = 5 \times 10^4 \).

\paragraph{Discussions.} Both Experience Replay (ER) and Elastic Weight Consolidation (EWC) demonstrate potential in mitigating catastrophic forgetting. However, they each come with notable limitations, particularly with respect to forward transfer performance, storage, and computational efficiency.

\textit{Storage Overhead:} ER demands significant storage space to maintain samples from prior tasks. This becomes particularly evident when comparing the storage needs of ER for larger datasets, such as the Kitchen dataset which requires 28GB, with the lightweight LoRA adapter occupies only 7.8MB. The vast difference in storage demands underscores the inefficiency of the ER approach.

\textit{Computational Challenges:} EWC, by design, necessitates the maintenance of a copy of the weights of the previous model in GPU memory. This leads to escalated GPU memory consumption, making EWC tends to reduce the training batch size, subsequently slowing down the training process.
    
\textit{Training Instability:} The regularization approach of EWC can introduce instability during training, owing to the regularization loss. This is also reflected by the poor forward transfer capability, as shown in Table \ref{tab:long-tasks-results}.

\textit{Scalability Concerns:} While EWC might be manageable for smaller networks, it is ill-suited for the fine-tuning of larger decision models due to its computational and storage challenges.

Given these outlined limitations, we advocate \method\ for alternative approaches that are both storage-efficient and computationally scalable, especially for large pretrained model adaptation.

\subsection{\method\ Adapter Configurations}
\label{sec:appendix:tail_adapters}

To establish our \method\ adapter configurations, we primarily draw from the AdapterHub implementation, setup and hyperparameters \citep{pfeiffer2020adapterhub}.

We utilize the default hyperparameters for LoRA, with the rank $r=8$ and scaling factor $\alpha=8$. These low-rank matrices are applied in parallel to the Transformer's query and value matrices \citep{hu2021lora}.
We also adopt the default for prefix token length of 30 for the prefix tuning \citep{li2021prefix} method across all tasks. To improve the training stability, Low-rank matrices ($r=16$) are employed during training to represent the prefix tokens.
The Bottleneck Adapter \citep{houlsby2019parameter} employs the bottleneck size of 32, and is applied to both the output layer of the attention and the intermediate feedforward layers.
The RoboAdapter method \citep{sharma2023lossless}, as the closest work to us, also applies the sequential adapters to the decision-making domain. It differs from the Bottleneck Adapter in that they adopt a special insertion of weights to specific layers of the Transformer, namely, layers $0,1,5,6,10,11$. They selectively skip certain layers, aiming to increase the bottleneck size on the remaining layers. Therefore, the bottleneck size is doubled to 64 for this approach, such that all methods share similar amount of parameters.

In order to maintain balanced adapter parameters number between the two CLIP-based (spatial and instruction) encoders, and the temporal transformer GPT2 decoder, the rank size for the GPT2 decoder is doubled across all methodologies. This adjustment compensates for the GPT2 decoder's fewer layers relative to the encoders.

For the continual learning setup, we use the previous stage's adapter weight (if any) plus a small random Gaussian noise with standard deviation 0.001 as an initialization of the current stage. The goal for adding a minor random noise aims to improve the adapter weight capacity \citep{kumar2022finetuning, agarwal2022reincarnating, lyle2022understanding}, preventing the weights from being trapped into local optimum. There is a potential to better utilize the trained adapter weights on preceding tasks. We outline several promising exploration directions in Appendix Section \ref{sec:appendix:future work}.

\subsection{Training Hyperparameters and Experiment Configurations}
\label{appendix:hyperparameter}
Following similar setup as in the LIBERO benchmark \citep{liu2023libero}, we perform data augmentation for the image observation data for all methods. We adopt the color, affine, and random erase augmentations to improve the robustness. The hyperparameters are presented in Table \ref{tab:training param}.

\begin{table}[h]
\centering
\caption{\small Image data augmentation and training hyperparameters}
\renewcommand{\arraystretch}{1.1}
\resizebox{1.\linewidth}{!}{
\begin{tabular}{ccccccccc}
\cline{1-4} \cline{6-9}
\multicolumn{4}{c}{\textbf{Image Augmentation}} & & \multicolumn{4}{c}{\textbf{Training and Optimizer Configuration}} \\
\cline{1-4} \cline{6-9}
Brightness & 0.3 & Contrast & 0.3 & & Training Epochs & 100/50 & Batch Size (per device) & 10/14/18 \\
Saturation & 0.3 & Hue & 0.3 & & Training Epochs per Eval & 5 & Eval Episodes/Task & 8 \\
Color Aug Prob. & 0.9 & Affine Degrees & 15 & & Warm-up Steps & 500 & Weight Decay & 0.1 \\
Affine Translate & 0.1 & Affine Prob. & 0.6 & & Learning Rate (LR) & 1e-4 & LR Scheduler & Linear \\
Random Erase Prob. & 0.1 & & & & Training Demo Num & 40 & Validation Demo Num & 40 \\
\cline{1-4} \cline{6-9}
\end{tabular}
}
\label{tab:training param}
\end{table}

For our training process, we employed the AdamW optimizer combined with a linear learning rate scheduler. The majority of our task suites—Kitchen, Spatial, Goal, Object, Living Room, and Study Room—underwent training for 100 epochs. Notably, each suite encompasses multiple tasks, with Kitchen having 40 and the others containing 8 each. In contrast, the 10 long-horizon adaptation tasks, termed LIBERO-10, were trained for 50 epochs, with each task trained sequentially.
We performed evaluations after every 5 training epochs over 8 episodes (unseen in training) for each task.

\textbf{Computing machine.}
Our experimental platform was powered by an AMD EPYC 7R32 CPU running Ubuntu 20.04.06. All trainings utilized 8 NVIDIA A10G GPUs, each with a memory of $22731$ MiB, equipped with driver version $470.199.02$ and CUDA version $11.4$. We employ Distributed Data Parallel (DDP) for parallel training across 8 GPUs, and utilize the 16-bit floating point precision (FP16) training mode to accelerate the training process.
To ensure reproducibility, we adopted 3 distinct random seeds: 0, 21, and 42.

\textbf{Training time.}
For a holistic perspective on training duration: FFT and ER methods demanded between $120\sim140$ hours per experiment ($1.5\sim1.75$ hours per task) for the 6 task suites shown in Fig. \ref{fig:training_curve_main}, including the evaluation time. In stark contrast, \method-based techniques slashed this to $60\sim66$ hours ($0.75\sim0.825$ hours per task). Hence, TAIL would also be much cheaper to train, considering its significantly shorter training time under identical computing machines.

Batch sizes varied by training method. EWC employed a batch size of 10, given its added memory demands to store a distinct full parameter set. FFT and ER utilized batch sizes of 14. Owing to \method's more efficient memory utilization—detailed in Table \ref{tab:parameters}—a larger batch size of 18 was feasible, which can maximize GPU resource usage on our machine, reducing training duration and cost.

\subsection{More Discussion and Future Directions}
\label{sec:appendix:future work}

The \method\ framework paves the way for a myriad of research opportunities:

\begin{enumerate}[leftmargin=*]
    \item \textbf{Better Weight Allocation Method Across Layers:} An interesting question within this framework is discerning which layers, early or later, derive the most benefit from weight modifications. This can offer insights into the adaptability of neural architectures \citep{lee2022surgical}.
    
    \item \textbf{Enhanced Reusability of Trained Adapters:} Exploring methods to efficiently reuse adapters from prior tasks, especially in scenarios with limited data, is a promising direction. AdapterFusion techniques \citep{pfeiffer2020adapterfusion} can be potentially useful, enabling the composition of knowledge from multiple pre-existing adapters.

    \item \textbf{Building on Knowledge with Parallel Integration:} The parallel integration method, particularly with LoRA weights, offers the capability to merge trained weights back into the main model. This iterative buildup of knowledge makes the approach valuable for continual learning, allowing new adapters to capitalize on the expertise of their predecessors.

    \item \textbf{Combining with Established Continual Learning Strategies:} The potential to merge the \method\ framework with existing continual learning methods, like Experience Replay and EWC, can be a beneficial avenue. Such integrations can accommodate the strengths of each method, crafting models that are both efficient in memory and robust against forgetting.

    \item \textbf{Extension beyond the Imitation Learning Domain:} Taking the \method framework into other decision-making domains, such as reinforcement learning (RL), is also promising. \method\ has the potential to address the model capacity loss issue in RL \citep{agarwal2022reincarnating, lyle2022understanding}. Leveraging the \method\ framework can also aid in multitask learning, meta-learning, and efficiently adapting offline-trained RL models to new tasks without the necessity of vast amounts of data or extensive fine-tuning, thereby potentially accelerating convergence to optimal policies.
\end{enumerate}

The avenues above elucidate the adaptability and potential of the \method\ framework, setting the stage for future research in this domain.

\clearpage
\section{More Experiment Results}
\label{sec:appendix:more_results}

\vspace{-2mm}
\subsection{Overfitting} 
\vspace{-2mm}
\label{sec:appendix:more_results:overfitting}
For each task, we used 40 demonstrations for training and 10 for validation. We are interested in the following question: \textit{In scenarios where data is limited, how resilient is \method\ against overfitting compared to traditional fine-tuning methods?} To answer this, we present the training and validation loss cross the Kitchen, Spatial, Goal, Object, Living Room and Study Room task suites, each with 100 epochs, in Fig. \ref{fig:loss_curve}.

\begin{figure}[h]
    \centering    \includegraphics[width=0.8\textwidth]{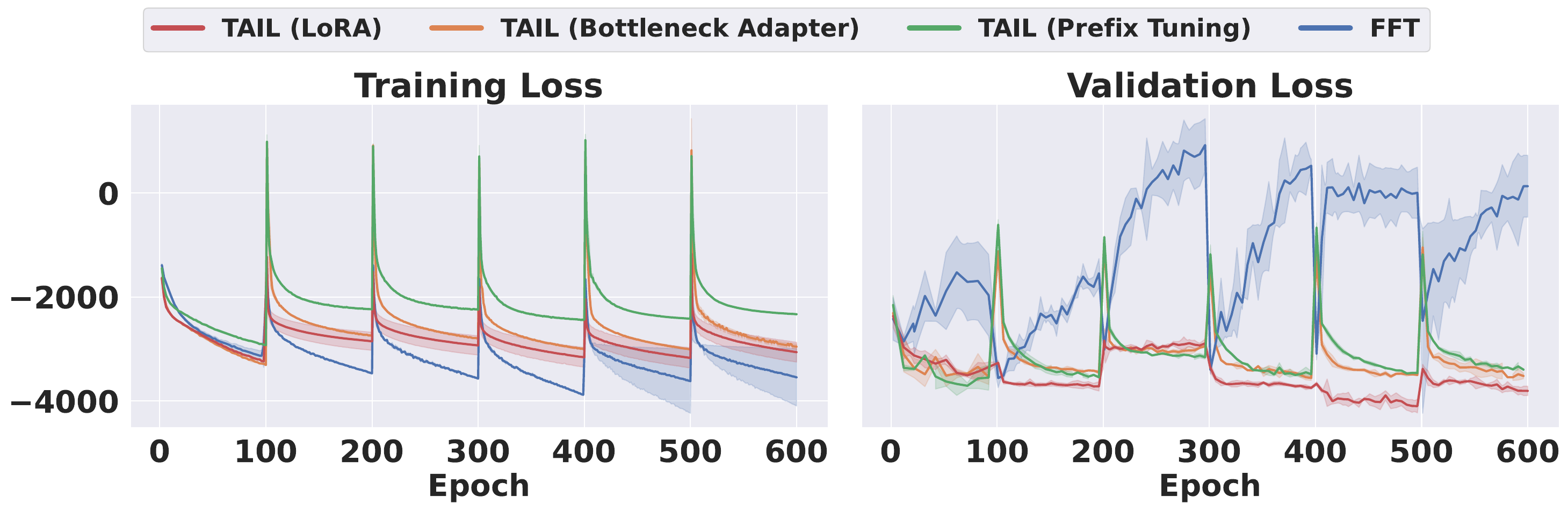}
        \caption{Adaptation loss trends: Training versus validation. The graph shows that the \method\ model consistently has more stable validation losses, which means that it is more robust to contexts with limited data. On the other hand, the full fine-tuning model (FFT) has larger validation losses, which means that it is more likely to overfit to the training data.}
        \label{fig:loss_curve}
\end{figure}

A noteworthy observation from Fig. \ref{fig:loss_curve} is the behavior of FFT. Despite achieving the lowest training loss across all stages, its validation loss spikes significantly after just a few epochs. This pattern suggests severe overfitting when FFT is applied to the entire parameter space using limited data. Intriguingly, this overfitting intensifies in the later adaptation phases, potentially signifying a distortion of pretrained features as alluded to by \citet{kumar2022finetuning}. Such distortion could be a contributor to the suboptimal success rate observed in Fig. \ref{fig:training_curve_main}, and the loss of learning capacity when revisiting a previous task, as presented in Table \ref{tab:circle back}.

\begin{wrapfigure}{R}{0.4\textwidth}
    \centering
    \vspace{-4mm}
    \includegraphics[width=0.99\linewidth]{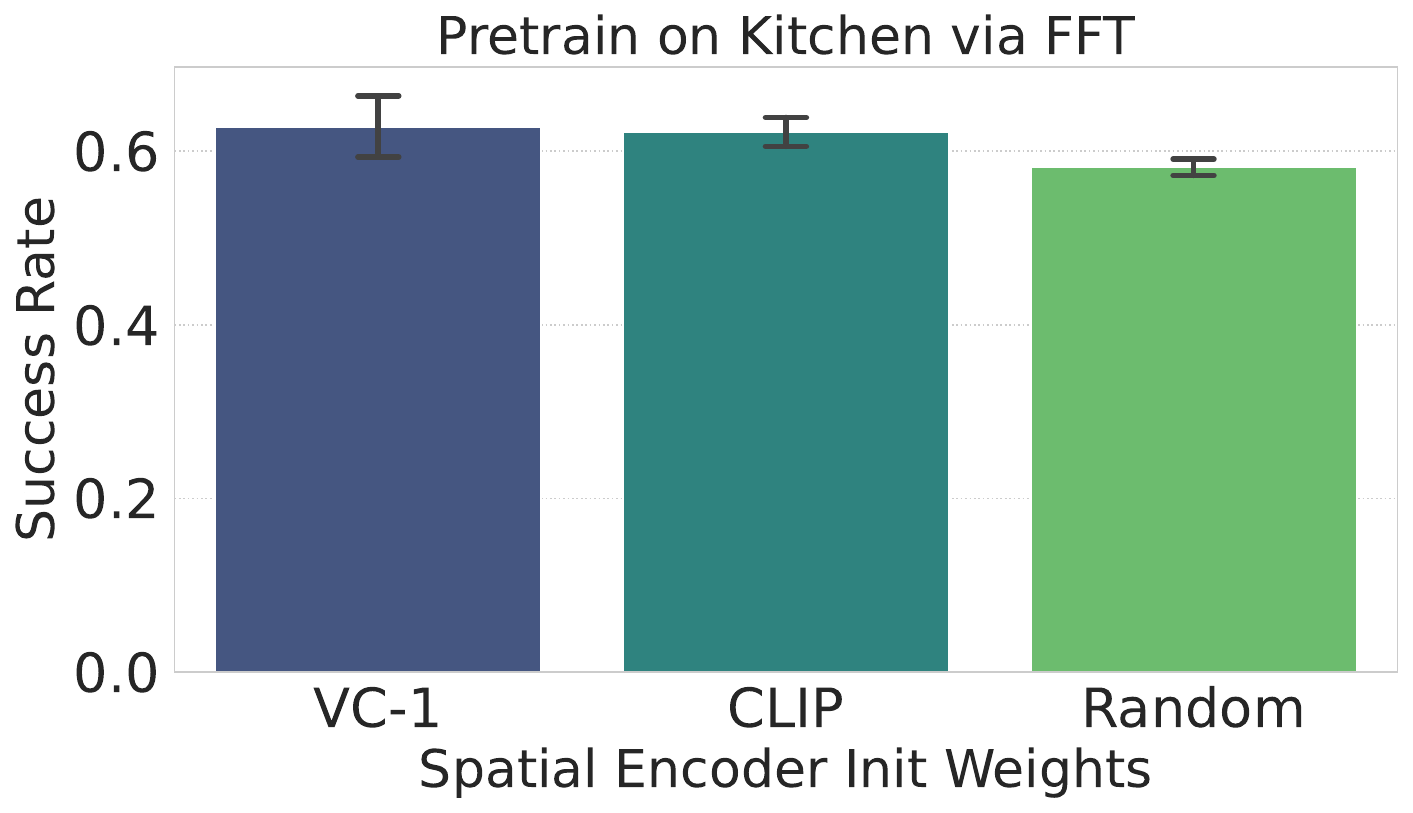}
    \caption{Training on the Kitchen task with different pretrained CLIP-ViT encoder weight. Random means using random initialization weight.}
    \label{fig:pretrain weight}
    \vspace{-2mm}
\end{wrapfigure}

In constrast, \method-based methods shows strong resilience against overfitting. Drawing from the Occam's razor principle, \method\ leverages fewer trainable parameters, inherently reducing its potential to overfit with scarce data. Additional, different integration styles provide the flexibility to effectively utilize the features from pretrained models while preserving them across all the adaptation stages.

This observation underscores the disparities between our decision-making problem, characterized by its limited data, and the traditional language or vision domains, which have data in abundance. Prior studies utilizing parameter-efficient fine-tuning techniques for language or vision tasks often reported superior performance with full fine-tuning due to its low training loss \citep{he2022towards, mao-etal-2022-unipelt, chen2022adaptformer, sharma2023lossless}. However, as our results demonstrate, a lower training loss does not invariably translate to superior performance, especially in the context of a data-scarce sequential decision-making tasks.

\subsection{Analysis of pretrained weights' influence} 
\label{sec:appendix:more_results:pretrained_influence}

We aim to answer the following question: \textit{how does the underlying pretrained base model influence the performance of \method, and are certain pretrained weights more conducive to this kind of adaptation?} 
We initiated our investigation by analyzing the success rates of 40 Kitchen tasks using different pretrained weights for the spatial encoder. Apart from the CLIP-ViT pretrained encodings as we adopted in our main results, two other initialization of weights were considered: one sourced from the Visual Cortex 1 (VC-1) \citep{majumdar2023we}, recognized for being a leading pretrained model for embodied agent tasks, and another using randomly initialized weights. The language instruction encoder consistently utilized the CLIP text model. From the results in Fig.~\ref{fig:pretrain weight}, the VC-1 pretrained weights delivered performance on par with the CLIP-ViT encodings. Both considerably outperformed the randomly initialized weights, suggesting that large-scale pretraining can indeed enhance downstream fine-tuning.
We then study how does the pretrained base model influence the performance of \method.

\subsection{Further Evaluations on \method\ with Different Base Models}
\label{sec:appendix:more_results:different_base_models}
To understand the influence of the base model's features on the performance of \method, we conducted additional evaluations. In Table \ref{tab:lora-pretrain-results}, the methods column showcases different configurations:

\begin{itemize}
    \item \textbf{LoRA (CLIP):} The main setup we adopted in the experiment section \ref{sec:exp}, which keeps the pretrained CLIP encodings frozen across all the adaptation stages.
    \item \textbf{LoRA (CLIP with FFT):} Starting with the CLIP model, we applied FFT pretraining on the Kitchen task before using LoRA for subsequent adaptations. \iclrnew{This helps us test out whether adaptation plasticity suffers after full fine-tuning as the only difference between this and the above method is the addition of full fine-tuning before using LoRA.}
    \item \textbf{LoRA (VC-1 with FFT):} The VC-1 model, after FFT pretraining on the Kitchen task, was adapted using LoRA.
    \item \textbf{LoRA (Random with FFT):} A model with randomly initialized weights underwent FFT pretraining on the Kitchen task, followed by adaptation with LoRA.
\end{itemize}

All the pretrained encodings implemented in the same model architecture as described in Appendix Section \ref{sec:appendix:architecture}.

Observations from Table \ref{tab:lora-pretrain-results} highlight several findings:

\begin{itemize}
    \item \textbf{Dominance of Original CLIP:} The pure CLIP base model, when combined with LoRA, yielded the highest success rates across all task suites, suggesting the inherent quality and robustness of the original CLIP features for these tasks.
    \item \textbf{FFT's Mixed Impact:} While FFT pretraining aids in task-specific fine-tuning, when combined with CLIP, it leads to a degradation in performance. This could be attributed to FFT potentially diluting the comprehensive and rich features within CLIP \iclrnew{while also reducing adaptation plasticity} \citep{kumar2022finetuning}, especially when exposed to a more constrained domain with limited data.
    \item \textbf{VC-1's Comparable Performance:} The VC-1 model, though renowned in the domain of embodied agent tasks, delivered results that were only marginally better than the randomly initialized weights when both were subjected to FFT pretraining and then adapted with LoRA. This emphasizes the unique advantages of the original CLIP features.
\end{itemize}

Interestingly, it is observed that CLIP is pretrained on the most comprehensive dataset, followed by VC-1. In contrast, the model with random weights only underwent pretraining on the 40 Kitchen tasks. The success rates mirror this order, underscoring the idea that the efficacy of \method\ is closely tied to a base model pretrained with rich features on extensive datasets.
So in summary, the choice of base model significantly affects the performance of \method, with CLIP's original features showing remarkable compatibility and resilience across various task suites

\begin{table}[h]
\caption{\small Evaluation results of FWT for LoRA with different pretrained model weights. The higher, the better.
We highlight the best method with highest FWT as \textbf{bold}.
}
\label{tab:lora-pretrain-results}
\large
\resizebox{1.\linewidth}{!}{
\begin{tabular}{|c|c|c|c|c|c|cl|}
\hline
Method            & Spatial                                            & Goal                                               & Object                                             & Living Room                                        & Study Room                                         & \multicolumn{2}{c|}{Average}                                            \\ \hline
LoRA (CLIP)       & \textbf{0.76 \scriptsize{± 0.02}} & \textbf{0.79 \scriptsize{± 0.02}} & \textbf{0.73 \scriptsize{± 0.14}} & \textbf{0.73 \scriptsize{± 0.07}} & \textbf{0.55 \scriptsize{± 0.11}} & \multicolumn{2}{c|}{\textbf{0.71 \scriptsize{± 0.07}}} \\
LoRA (CLIP with FFT)   & 0.62 \scriptsize{± 0.04}          & 0.67 \scriptsize{± 0.13}          & 0.38 \scriptsize{± 0.08}          & 0.32 \scriptsize{± 0.08}          & 0.32 \scriptsize{± 0.01}          & \multicolumn{2}{c|}{0.46 \scriptsize{± 0.07}}          \\
LoRA (Random with FFT) & 0.38 \scriptsize{± 0.19}          & 0.60 \scriptsize{± 0.06} & 0.37 \scriptsize{± 0.03}          & 0.23 \scriptsize{± 0.01}          & 0.47 \scriptsize{± nan}           & \multicolumn{2}{c|}{0.41 \scriptsize{± 0.07}}          \\
LoRA (VC-1 with FFT)   & 0.56 \scriptsize{± 0.07}          & 0.66 \scriptsize{± 0.08}          & 0.25 \scriptsize{± 0.00}          & 0.20 \scriptsize{± 0.06}          & 0.48 \scriptsize{± 0.07}          & \multicolumn{2}{c|}{0.43 \scriptsize{± 0.05}}          \\ \hline
\end{tabular}
}
\end{table}

\iclrnew{
\subsection{Rank Size Ablation Study} 
\label{sec:appendix:more_results:rank_ablation}
In order to understand the impact of rank-size on adaptation performance, we conducted experiments using varying rank sizes for the LoRA and Bottleneck Adapter methods. The results, illustrated in Fig. \ref{fig:rank_ablation}, present the average success rates across the Spatial, Goal, and Object task suites. It is evident that increasing the rank size generally enhances performance up to a certain point. Beyond this optimal threshold, further increasing the rank size does not necessarily lead to higher success rates, potentially because of overfitting. Notably, in our continual learning context, the parallel insertion approach of LoRA consistently surpasses the sequential style of the Bottleneck Adapter method.}

\begin{figure}[ht]
    \centering
    \begin{minipage}{0.45\textwidth}
        \centering
        \includegraphics[width=0.99\linewidth]{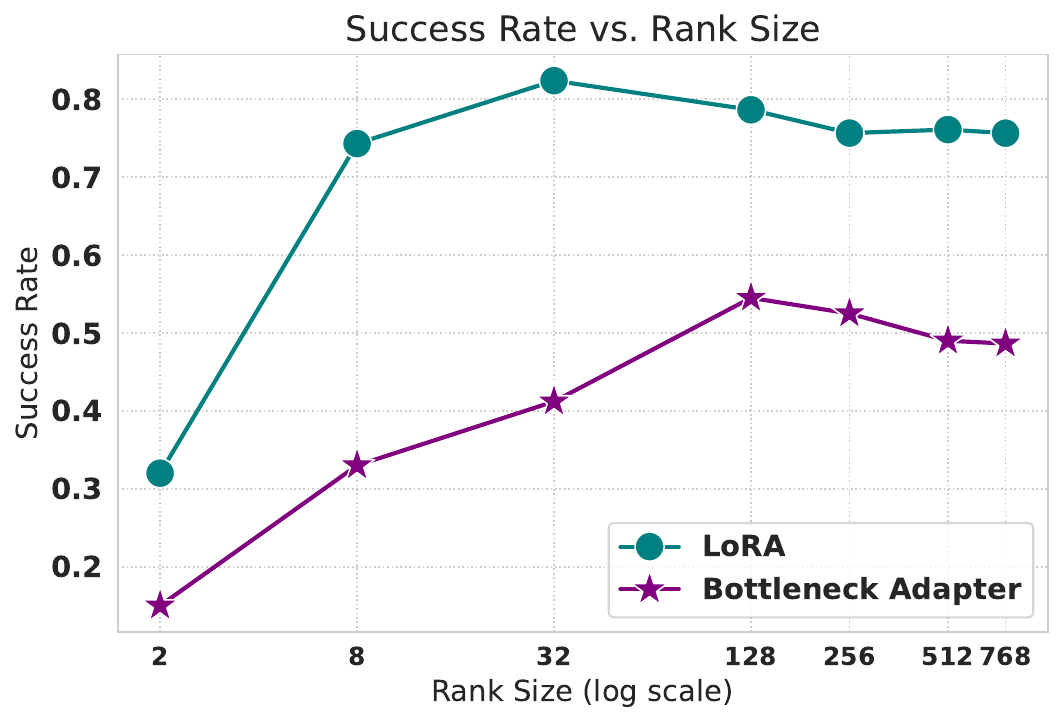}
        \caption{\small Ablation study of the rank-size of LoRA and Bottleneck adapters. Increasing the rank size generally enhances performance up to a certain point. Beyond this optimal threshold, further increasing the rank size does not necessarily lead to higher success rates. The parallel insertion approach of LoRA consistently surpasses the sequential style of the Bottleneck Adapter method}
        \label{fig:rank_ablation}
    \end{minipage}\hfill
    \begin{minipage}{0.5\textwidth}
        \centering
        \includegraphics[width=0.99\linewidth]{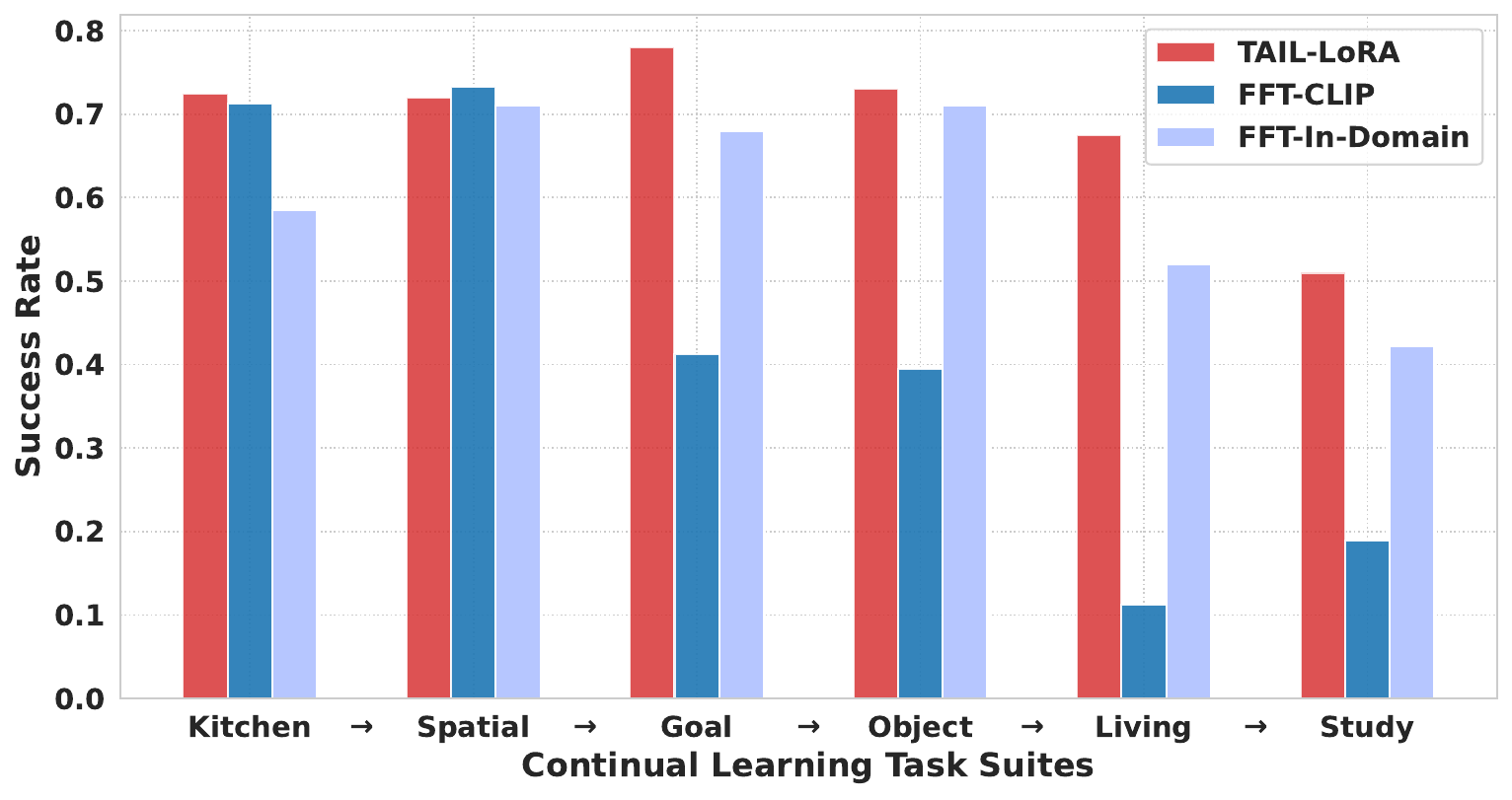}
        \caption{\small Comparison for TAIL-LoRA, sequential FFT with pre-trained CLIP weights, and FFT-In-Domain. FFT-In-Domain is trained from scratch with task-suite demonstration data only, which saves a copy of the entire model for each task. FFT with CLIP excels in initial Kitchen and Spatial suites, highlighting the value of pretrained models; however, its performance declines in subsequent tasks, suggesting reduced adaptability. In contrast, TAIL-LoRA demonstrates consistent superior performance across all suites.}
        \label{fig:scratch-baseline}
    \end{minipage}
\end{figure}

\iclrnew{
Additionally, we would like to note that our TAIL framework exhibits data adaptivity, suggesting that the rank size could be adjusted based on the quantity of adaptation data. In scenarios with smaller datasets, a smaller rank size could be more effective, and vice versa. 
}

\vspace{-3mm}
\iclrnew{
\subsection{Comparison between Training from Scratch and Using Pretrained Models}
\label{sec:appendix:more_results:train_from_scratch}
Fig. \ref{fig:scratch-baseline} compares the success rates across task suites for TAIL-LoRA, sequential FFT with pre-trained CLIP weights, and FFT-In-Domain. Unlike FFT-CLIP, FFT-In-Domain is trained from scratch with task-suite demonstration data only, i.e., we need to maintain a copy of the entire model for each task suite. There are three observations:}

\iclrnew{
\textbf{1. Pretrained Weights Advantage:} In the initial Kitchen and Spatial task suites, FFT with CLIP pretrained weights demonstrates a higher success rate compared to FFT trained from scratch. This indicates the effectiveness of leveraging pretrained models, particularly in the context of the Kitchen suite where the benefit is more pronounced.}

\iclrnew{
\textbf{2. Decline in Model Adaptability:} Despite the initial advantage, sequential FFT with CLIP shows a marked decline in performance in the remaining four task suites - Goal, Object, Living, and Study. This trend may be indicative of a loss in model plasticity, where the pre-trained model performs well in the early stages but struggles to adapt to new tasks after the pre-trained weights are contaminated.}

\iclrnew{
\textbf{3. TAIL-LoRA's Consistent Performance:} Throughout all the task suites, TAIL-LoRA with pretrained CLIP consistently outperforms the other methods. This suggests that the LoRA approach, combined with the advantages of pretrained CLIP weights, provides a robust and adaptable framework capable of handling a variety of tasks with greater efficiency.}

\iclrnew{
\subsection{Ablation study for different integration style combinations}}
\label{sec:appendix:more_results:combination}

\iclrnew{It's noteworthy that our method allows for the simultaneous use of multiple integration techniques \citep{mao-etal-2022-unipelt}. This flexibility lets us explore the performance impact of combining LoRA (parallel integration), bottleneck adapter (sequential integration), and prefix token (concatenation). To this end, we conduct an ablation study for each of the combinations over the Spatial, Goal, and Object task suites. The experiment result is shown in Fig. \ref{fig:combination}, where the y-axis is the averaged success rate.}

\begin{wrapfigure}{R}{0.6\textwidth}
    \centering
    \vspace{-2mm}
    \includegraphics[width=0.99\linewidth]{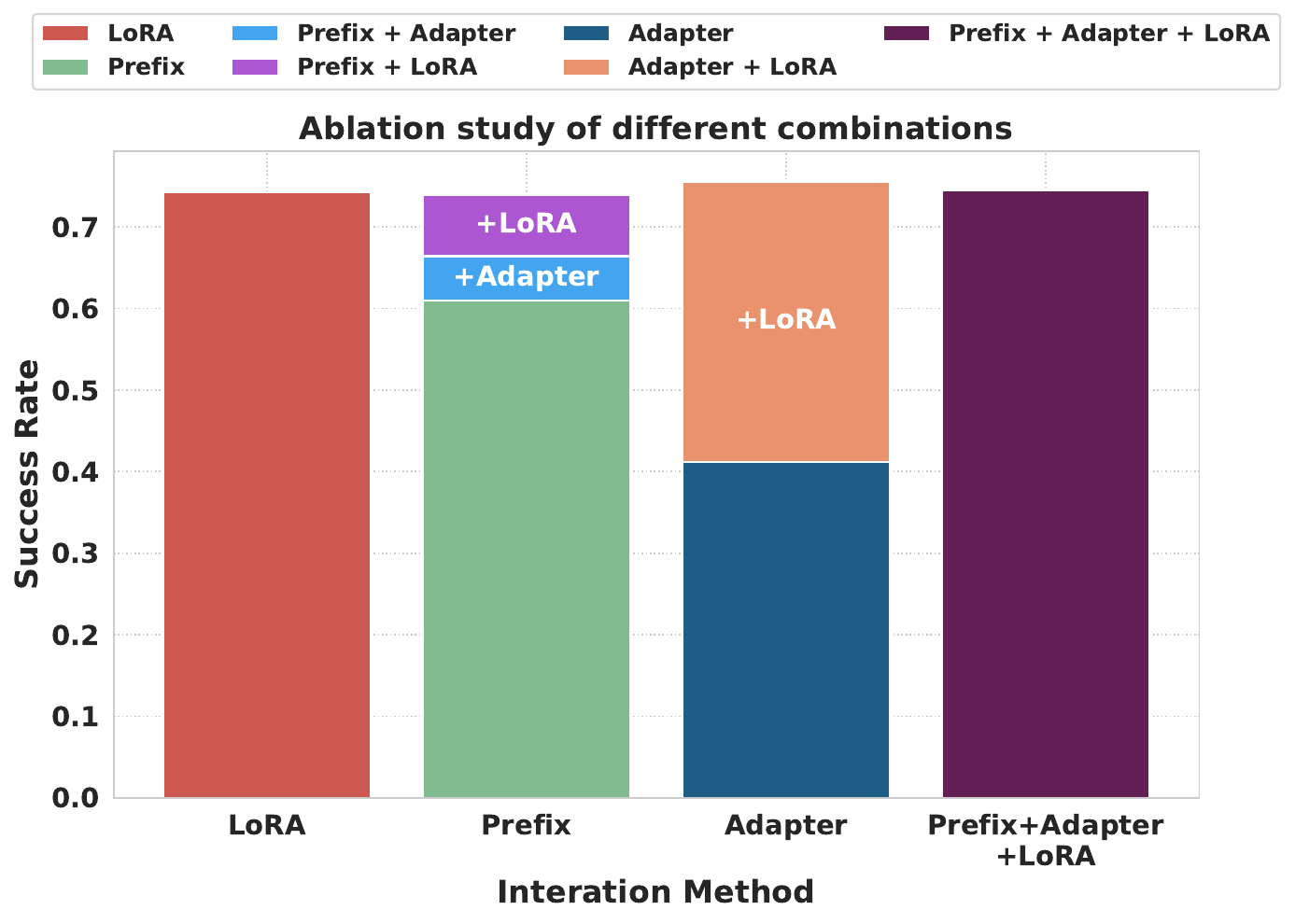}
    \caption{\small Ablation study for integration style combinations. LoRA (parallel integration) plays a crucial role in enhancing adaptation performance, consistently outperforming methods without it. Whether used alone or in combination with other methods like Prefix and Adapter, LoRA shows superior effectiveness.}
    \label{fig:combination}
    \vspace{-2mm}
\end{wrapfigure}

\iclrnew{A key finding is the critical role of LoRA (parallel integration) in enhancing adaptation performance. Combinations involving LoRA consistently outperform those without it. For instance, the standalone use of LoRA yields a comparable success rate w.r.t the combination with others. This pattern underscores LoRA's effectiveness, either used alone or in conjunction with other methods. In contrast, the combination of Prefix and Adapter without LoRA results in a notably lower success rate (0.6641), highlighting LoRA's indispensability.}

\iclrnew{The integration of all three methods—Prefix, Adapter, and LoRA—achieves a success rate that is comparable to LoRA's standalone performance. This outcome suggests that while the combination of different integration methods does not detract from performance, LoRA remains the primary driver of successful adaptation. These findings emphasize the importance of LoRA in adapter weight integration strategies and provide valuable guidance for future approaches in this domain.}

\subsection{Detailed per-task results in the LIBERO-Long task suite}
\begin{table}[!h]
    \centering
    \caption{\small Adaptation results on 10 long horizon tasks. The $\uparrow$ symbol means the higher, the better. The BWT $\uparrow$ for TAIL methods are all 0 (no catastrophic forgetting). We highlight the best method (highest FWT $\uparrow$) in \textbf{bold}. FPF results were omitted due to its near-zero performance.}
    \label{tab:long-tasks-results-appendix}
    \small
    \renewcommand{\arraystretch}{1.1}
    \resizebox{\linewidth}{!}{
        \begin{tabular}{@{}lcccccc>{\columncolor[gray]{0.9}}c>{\columncolor[gray]{0.9}}c>{\columncolor[gray]{0.9}}c>{\columncolor[gray]{0.9}}c>{\columncolor[gray]{0.9}}c@{}}
            \toprule
            \multirow{3}{*}{Task} & \multicolumn{6}{c}{Conventional Fine-Tuning Methods} & \multicolumn{4}{c}{TAIL-based Methods (\textbf{Ours})} \\ \cmidrule(lr){2-7} \cmidrule(lr){8-11}
            & \multicolumn{2}{c}{Full Fine-Tuning} & \multicolumn{2}{c}{Experience Replay} & \multicolumn{2}{c}{EWC} & \multicolumn{1}{c}{LoRA} & \multicolumn{1}{c}{Prefix} & \multicolumn{1}{c}{Bottleneck} & \multicolumn{1}{c}{RoboAdapter} \\
            & FWT $\uparrow$ & BWT $\uparrow$ & FWT $\uparrow$ & BWT $\uparrow$ & FWT $\uparrow$ & BWT $\uparrow$ & \multicolumn{1}{c}{FWT $\uparrow$} & \multicolumn{1}{c}{FWT $\uparrow$} & \multicolumn{1}{c}{FWT $\uparrow$} & \multicolumn{1}{c}{FWT $\uparrow$} \\ \midrule
            Task 1 & 0.42 \scriptsize$\pm$ 0.07 & - & 0.25 \scriptsize$\pm$ 0.12 & - & 0.38 \scriptsize$\pm$ 0.12 & - & \textbf{0.62 \scriptsize$\pm$ 0.00} & 0.38 \scriptsize$\pm$ 0.12 & 0.21 \scriptsize$\pm$ 0.14 & 0.12 \scriptsize$\pm$ 0.00 \\
            Task 2 & 0.58 \scriptsize$\pm$ 0.07 & -0.42 \scriptsize$\pm$ 0.06 & 0.58 \scriptsize$\pm$ 0.07 & -0.25 \scriptsize$\pm$ 0.10 & 0.54 \scriptsize$\pm$ 0.07 & -0.38 \scriptsize$\pm$ 0.10 & \textbf{0.75 \scriptsize$\pm$ 0.00} & 0.58 \scriptsize$\pm$ 0.19 & \textbf{0.75 \scriptsize$\pm$ 0.12} & 0.50 \scriptsize$\pm$ 0.12 \\
            Task 3 & 0.71 \scriptsize$\pm$ 0.07 & -0.50 \scriptsize$\pm$ 0.10 & 0.67 \scriptsize$\pm$ 0.07 & -0.42 \scriptsize$\pm$ 0.19 & 0.38 \scriptsize$\pm$ 0.12 & -0.46 \scriptsize$\pm$ 0.12 & \textbf{0.96 \scriptsize$\pm$ 0.07} & 0.88 \scriptsize$\pm$ 0.22 & 0.71 \scriptsize$\pm$ 0.19 & 0.50 \scriptsize$\pm$ 0.25 \\
            Task 4 & \textbf{0.96 \scriptsize$\pm$ 0.07} & -0.57 \scriptsize$\pm$ 0.13 & 0.92 \scriptsize$\pm$ 0.07 & -0.50 \scriptsize$\pm$ 0.20 & 0.75 \scriptsize$\pm$ 0.25 & -0.43 \scriptsize$\pm$ 0.12 & 0.88 \scriptsize$\pm$ 0.00 & 0.71 \scriptsize$\pm$ 0.07 & 0.71 \scriptsize$\pm$ 0.19 & 0.58 \scriptsize$\pm$ 0.14 \\
            Task 5 & 0.21 \scriptsize$\pm$ 0.07 & -0.67 \scriptsize$\pm$ 0.21 & 0.33 \scriptsize$\pm$ 0.14 & -0.60 \scriptsize$\pm$ 0.25 & 0.17 \scriptsize$\pm$ 0.19 & -0.50 \scriptsize$\pm$ 0.18 & \textbf{0.62 \scriptsize$\pm$ 0.12} & 0.17 \scriptsize$\pm$ 0.07 & 0.25 \scriptsize$\pm$ 0.00 & 0.29 \scriptsize$\pm$ 0.07 \\
            Task 6 & \textbf{0.83 \scriptsize$\pm$ 0.19} & -0.57 \scriptsize$\pm$ 0.26 & 0.71 \scriptsize$\pm$ 0.19 & -0.55 \scriptsize$\pm$ 0.25 & 0.50 \scriptsize$\pm$ 0.43 & -0.42 \scriptsize$\pm$ 0.19 & 0.75 \scriptsize$\pm$ 0.12 & 0.79 \scriptsize$\pm$ 0.14 & 0.75 \scriptsize$\pm$ 0.00 & 0.75 \scriptsize$\pm$ 0.25 \\
            Task 7 & 0.17 \scriptsize$\pm$ 0.07 & -0.62 \scriptsize$\pm$ 0.27 & 0.12 \scriptsize$\pm$ 0.00 & -0.58 \scriptsize$\pm$ 0.25 & 0.04 \scriptsize$\pm$ 0.07 & -0.44 \scriptsize$\pm$ 0.24 & \textbf{0.54 \scriptsize$\pm$ 0.26} & 0.38 \scriptsize$\pm$ 0.12 & 0.31 \scriptsize$\pm$ 0.09 & 0.33 \scriptsize$\pm$ 0.07 \\
            Task 8 & 0.42 \scriptsize$\pm$ 0.07 & -0.55 \scriptsize$\pm$ 0.29 & 0.29 \scriptsize$\pm$ 0.07 & -0.51 \scriptsize$\pm$ 0.28 & 0.12 \scriptsize$\pm$ 0.18 & -0.46 \scriptsize$\pm$ 0.28 & \textbf{0.75 \scriptsize$\pm$ 0.25} & 0.67 \scriptsize$\pm$ 0.19 & 0.25 \scriptsize$\pm$ 0.18 & 0.50 \scriptsize$\pm$ 0.22 \\
            Task 9 & 0.17 \scriptsize$\pm$ 0.07 & -0.54 \scriptsize$\pm$ 0.28 & 0.12 \scriptsize$\pm$ 0.05 & -0.50 \scriptsize$\pm$ 0.28 & 0.00 \scriptsize$\pm$ 0.00 & -0.41 \scriptsize$\pm$ 0.29 & \textbf{0.38 \scriptsize$\pm$ 0.12} & 0.08 \scriptsize$\pm$ 0.07 & 0.19 \scriptsize$\pm$ 0.09 & 0.21 \scriptsize$\pm$ 0.07 \\
            Task 10 & 0.33 \scriptsize$\pm$ 0.19 & -0.50 \scriptsize$\pm$ 0.29 & 0.50 \scriptsize$\pm$ 0.02 & -0.46 \scriptsize$\pm$ 0.29 & 0.12 \scriptsize$\pm$ 0.18 & -0.38 \scriptsize$\pm$ 0.31 & \textbf{0.79 \scriptsize$\pm$ 0.07} & 0.50 \scriptsize$\pm$ 0.33 & 0.44 \scriptsize$\pm$ 0.09 & 0.42 \scriptsize$\pm$ 0.07 \\ \midrule
            Average & 0.48 \scriptsize$\pm$ 0.10 & -0.55 \scriptsize$\pm$ 0.21 & 0.45 \scriptsize$\pm$ 0.09 & -0.49 \scriptsize$\pm$ 0.23 & 0.30 \scriptsize$\pm$ 0.16 & -0.43 \scriptsize$\pm$ 0.20 & \textbf{0.70 \scriptsize$\pm$ 0.10} & 0.51 \scriptsize$\pm$ 0.15 & 0.46 \scriptsize$\pm$ 0.11 & 0.42 \scriptsize$\pm$ 0.13 \\ \bottomrule
        \end{tabular}
    }
\end{table}

\section{Evaluation Task Details}
\label{sec:appendix:task details}

We list all the language instructions describing the tasks we adopted in our experiments below.
\iclrnew{Note that while certain tasks may share similar descriptions, they are not the same due to variations in the environment configurations (e.g., different spatial layouts, objects, or goal positions).}

\begin{table}[ht]
    \centering
    \begin{tabularx}{\linewidth}{|l|X|}
        \hline
        \textbf{Task Suite} & \textbf{Instructions} \\
        \hline
        & close the top drawer of the cabinet \\
        & close the top drawer of the cabinet and put the black bowl on top of it \\
        Kitchen  & put the black bowl in the top drawer of the cabinet \\
        & put the butter at the back in the top drawer of the cabinet and close it \\
        & put the butter at the front in the top drawer of the cabinet and close it \\
        & put the chocolate pudding in the top drawer of the cabinet and close it \\
        & open the bottom drawer of the cabinet \\
        & open the top drawer of the cabinet \\
        & open the top drawer of the cabinet and put the bowl in it \\
        & put the black bowl on the plate \\
        & put the black bowl on top of the cabinet \\
        & open the top drawer of the cabinet \\
        & put the black bowl at the back on the plate \\
        & put the black bowl at the front on the plate \\
        & put the middle black bowl on the plate \\
        & put the middle black bowl on top of the cabinet \\
        & stack the black bowl at the front on the black bowl in the middle \\
        & stack the middle black bowl on the back black bowl \\
        & put the frying pan on the stove \\
        & put the moka pot on the stove \\
        & turn on the stove \\
        & turn on the stove and put the frying pan on it \\
        & close the bottom drawer of the cabinet \\
        & close the bottom drawer of the cabinet and open the top drawer \\
        & put the black bowl in the bottom drawer of the cabinet \\
        & put the black bowl on top of the cabinet \\
        & put the wine bottle in the bottom drawer of the cabinet \\
        & put the wine bottle on the wine rack \\
        & close the top drawer of the cabinet \\
        & put the black bowl in the top drawer of the cabinet \\
        & put the black bowl on the plate \\
        & put the black bowl on top of the cabinet \\
        & put the ketchup in the top drawer of the cabinet \\
        & close the microwave \\
        & put the yellow and white mug to the front of the white mug \\
        & open the microwave \\
        & put the white bowl on the plate \\
        & put the white bowl to the right of the plate \\
        & put the right moka pot on the stove \\
        & turn off the stove \\
        \hline
    \end{tabularx}
    \caption{40 Kitchen scene pretraining tasks}
\end{table}

\begin{table}[ht]
    \centering
    \begin{tabularx}{\linewidth}{|l|X|}
        \hline
        \textbf{Task Suite} & \textbf{Instructions} \\
        \hline
         & put both the alphabet soup and the tomato sauce in the basket \\
        Long-horizon & put both the cream cheese box and the butter in the basket \\
        (LIBERO 10) & turn on the stove and put the moka pot on it \\
        & put the black bowl in the bottom drawer of the cabinet and close it \\
        & put the white mug on the left plate and put the yellow and white mug on the right plate \\
        & pick up the book and place it in the back compartment of the caddy \\
        & put the white mug on the plate and put the chocolate pudding to the right of the plate \\
        & put both the alphabet soup and the cream cheese box in the basket \\
        & put both moka pots on the stove \\
        & put the yellow and white mug in the microwave and close it \\
        \hline
        & pick up the black bowl between the plate and the ramekin and place it on the plate \\
       Spatial  & pick up the black bowl next to the ramekin and place it on the plate \\
        & pick up the black bowl from table center and place it on the plate \\
        & pick up the black bowl on the cookie box and place it on the plate \\
        & pick up the black bowl in the top drawer of the wooden cabinet and place it on the plate \\
        & pick up the black bowl on the ramekin and place it on the plate \\
        & pick up the black bowl next to the cookie box and place it on the plate \\
        & pick up the black bowl on the stove and place it on the plate \\

        \hline
        & open the middle drawer of the cabinet \\
       Goal  & put the bowl on the stove \\
        & put the wine bottle on top of the cabinet \\
        & open the top drawer and put the bowl inside \\
        & put the bowl on top of the cabinet \\
        & push the plate to the front of the stove \\
        & put the cream cheese in the bowl \\
        & turn on the stove \\
        \hline
        & pick up the alphabet soup and place it in the basket \\
        Object & pick up the cream cheese and place it in the basket \\
        & pick up the salad dressing and place it in the basket \\
        & pick up the bbq sauce and place it in the basket \\
        & pick up the ketchup and place it in the basket \\
        & pick up the tomato sauce and place it in the basket \\
        & pick up the butter and place it in the basket \\
        & pick up the milk and place it in the basket \\
        \hline
        & pick up the alphabet soup and put it in the basket \\
       Living Room  & pick up the butter and put it in the basket \\
        & pick up the milk and put it in the basket \\
        & pick up the orange juice and put it in the basket \\
        & pick up the tomato sauce and put it in the basket \\
        & pick up the alphabet soup and put it in the tray \\
        & pick up the butter and put it in the tray \\
        & pick up the cream cheese and put it in the tray \\
        \hline
        & pick up the book and place it in the right compartment of the caddy \\
       Study Room & pick up the book and place it in the front compartment of the caddy \\
        & pick up the book and place it in the left compartment of the caddy \\
        & pick up the book and place it in the right compartment of the caddy \\
        & pick up the red mug and place it to the right of the caddy \\
        & pick up the white mug and place it to the right of the caddy \\
        & pick up the book in the middle and place it on the cabinet shelf \\
        & pick up the book on the left and place it on top of the shelf \\
        \hline
    \end{tabularx}
    \caption{Adaptation task suites}
\end{table}

\end{document}